\documentclass[letterpaper, 10 pt, conference]{ieeeconf}
\pdfoutput=1
\IEEEoverridecommandlockouts
\usepackage[english]{babel}
\usepackage[utf8]{inputenc}
\usepackage{latexsym}
\usepackage{enumerate}
\usepackage{ifthen}
\usepackage{subfiles}
\usepackage{import}
\usepackage{ifmtarg}
\usepackage{xparse}
\usepackage{float}
\usepackage{color}
\usepackage{setspace}
\usepackage[%
    colorlinks=true,
    bookmarksopen,
    bookmarksnumbered,
    allcolors=black
]{hyperref}

\usepackage{tabularx}
\usepackage{array}
\usepackage{multirow}
\usepackage{multicol}
\usepackage{booktabs}
\usepackage{threeparttable}
\usepackage{amsfonts, bbold, yfonts, dsfont}
\usepackage{xcolor}
\usepackage{graphicx, wrapfig, epstopdf}
\graphicspath{{./images/}}

\usepackage{afterpage}
\usepackage[font={small,sf},justification=centering]{caption}
\usepackage{subcaption}
\usepackage{mathtools, mathrsfs, nicefrac}
\usepackage{breqn, amssymb, amscd}
\usepackage{textcomp}
\usepackage{gensymb}
\usepackage{amsmath}
\usepackage{algorithm2e}

\SetCommentSty{algCommentFont}
\SetKwComment{Comment}{/* }{ */}
\SetKwInput{KwInput}{Input}                
\SetKwInput{KwOutput}{Output}              
\RestyleAlgo{ruled}
\DontPrintSemicolon
\LinesNumbered
\SetSideCommentLeft
\newcommand{\parName}[1]{\medskip \noindent #1 \noindent}
\newcommand{\mboxTiny}[1]{\makebox[0pt]{\mbox{\normalfont\tiny\sffamily #1}}}

\newcommand{\assign}[2]{
\ifdefined#1
\renewcommand{#1}{#2}
\else
\newcommand{#1}{#2}
\fi
}
\newcommand{\unassign}[1]{
\ifdefined#1
\let#1\undefined
\fi
}
\newcommand{\DefineMathCmd}[2]{\newcommand{#1}{\ensuremath{#2}}}

\title{\LARGE \bf Active perception network for non-myopic online exploration and visual surface coverage}
\author{David Vutetakis$^{1}$ and Jing Xiao$^{2}$
\thanks{$^{1}$David Vutetakis is with the Department of Computer Science, University of North Carolina, Charlotte, NC 28223, USA.{\tt\small dvutetak@uncc.edu}}%
\thanks{$^{2}$Jing Xiao is with the Department of Robotics Engineering, Worcester Polytechnic Institute, Worcester, MA 01609, USA. {\tt\small jxiao2@wpi.edu}}%
}

\begin{document}

\maketitle
\thispagestyle{empty}
\pagestyle{empty}

\begin{abstract}
This work addresses the problem of online exploration and visual sensor coverage of unknown environments. We introduce a novel perception roadmap we refer to as the Active Perception Network (APN) that serves as a hierarchical topological graph describing how to traverse and perceive an incrementally built spatial map of the environment. The APN state is incrementally updated to expand a connected configuration space that extends throughout as much of the known space as possible, using efficient difference-awareness techniques that track the discrete changes of the spatial map to inform the updates. A frontier-guided approach is presented for efficient evaluation of information gain and covisible information, which guides view sampling and refinement to ensure maximum coverage of the unmapped space is maintained within the APN. The updated roadmap is hierarchically decomposed into subgraph regions which we use to facilitate a non-myopic global view sequence planner. A comparative analysis to several state-of-the-art approaches was conducted, showing significant performance improvements in terms of total exploration time and surface coverage, and demonstrating high computational efficiency that is scalable to large and complex environments.
\end{abstract}


\DefineMathCmd{\symReals}{\mathbb{R}}
\DefineMathCmd{\symNaturals}{\mathbb{N}}
\DefineMathCmd{\symIntegers}{\mathbb{Z}}
\DefineMathCmd{\symPosIntegers}{\symIntegers^{+}}
\DefineMathCmd{\symNegIntegers}{\symIntegers^{-}}
\DefineMathCmd{\symCardinals}{\aleph}
\DefineMathCmd{\symComplexGroup}{\mathbb{C}}
\DefineMathCmd{\symOrthoGroup}{O}
\DefineMathCmd{\symSEGroup}{SE}
\DefineMathCmd{\symSOGroup}{SO}
\DefineMathCmd{\symProjGroup}{\mathbb{P}}
\DefineMathCmd{\symProjPlaneGroup}{\symReals\symProjGroup^{2}}
\DefineMathCmd{\symPowerSet}{\mathcal{P}}
\DefineMathCmd{\IndicatorFn}{\mathbb{1}}
\DefineMathCmd{\unitsMeters}{\text{m}}
\DefineMathCmd{\unitsMetersSq}{\text{m}^2}
\DefineMathCmd{\unitsMetersCubed}{\text{m}^3}
\DefineMathCmd{\unitsSeconds}{\text{s}}
\DefineMathCmd{\unitsSecondsSq}{\text{s}^2}
\DefineMathCmd{\unitsMillisec}{\text{ms}}
\DefineMathCmd{\unitsVelocity}{\text{m/s}}
\DefineMathCmd{\unitsAngVelocity}{\text{rad/s}}
\DefineMathCmd{\Sec}{\text{s}}
\DefineMathCmd{\mSec}{\text{ms}}
\DefineMathCmd{\Meters}{\text{m}}
\DefineMathCmd{\SqMeters}{\text{m}^2}
\DefineMathCmd{\CubicMeters}{\text{m}^3}
\DefineMathCmd{\CubicMetersPerSec}{\text{m}^{3}/\text{s}}
\DefineMathCmd{\unitsHz}{\text{Hz}}

\DefineMathCmd{\symCoord}{\boldsymbol{x}}
\DefineMathCmd{\symPos}{\boldsymbol{x}}
\DefineMathCmd{\symPoint}{\boldsymbol{x}}
\DefineMathCmd{\symPnt}{\symPoint}
\DefineMathCmd{\symEulerVec}{\boldsymbol{a}}
\DefineMathCmd{\symQuat}{\boldsymbol{h}}
\DefineMathCmd{\symPose}{\boldsymbol{q}}
\DefineMathCmd{\symPoseTransform}{\boldsymbol{T}}
\DefineMathCmd{\symRobotPose}{\symPose^{agent}}
\DefineMathCmd{\symRobotPosn}{\symPos^{agent}}
\DefineMathCmd{\symRobotQuat}{\symQuat^{agent}}
\DefineMathCmd{\symPosePath}{\Xi}
\DefineMathCmd{\symRoll}{\varphi}
\DefineMathCmd{\symPitch}{\vartheta}
\DefineMathCmd{\symYaw}{\psi}
\DefineMathCmd{\symBoundingVol}{B}
\DefineMathCmd{\symWSpace}{\ensuremath{\mathcal{W}}}
\DefineMathCmd{\symWSpaceOcc}{\ensuremath{\symWSpace_{occ}}}
\DefineMathCmd{\symWSpaceFree}{\ensuremath{\symWSpace_{free}}}
\DefineMathCmd{\symWSpaceUnk}{\ensuremath{\symWSpace_{unk}}}
\DefineMathCmd{\symCSpace}{\mathcal{X}}
\DefineMathCmd{\symReachSpace}{\symCSpace}
\DefineMathCmd{\symReachCost}{L}
\DefineMathCmd{\symSurfSpace}{\mathcal{S}}
\DefineMathCmd{\symVisibleSurfs}{\symSurfSpace_{\symCSpace}}
\DefineMathCmd{\symNonvisSurfs}{\symSurfSpace_{nonvis}}
\DefineMathCmd{\symResidualSurfs}{ \symSurfSpace_{res} }

\DefineMathCmd{\symVisSpace}{\mathcal{M}_{\Theta}}
\DefineMathCmd{\symXStateSpace}{\Omega}
\DefineMathCmd{\symCollisionState}{l^{\mathcal{O}}}
\DefineMathCmd{\symCollSpace}{\symCSpace^{\mathcal{O}}}


\DefineMathCmd{\symMaxVel}{\boldsymbol{v}_{max}}
\DefineMathCmd{\symMaxYawRate}{\dot{\symYaw}_{max}}
\DefineMathCmd{\symMaxYaw}{\symMaxYawRate}
\DefineMathCmd{\symSafetyVol}{\symBoundingVol^{safe}}
\DefineMathCmd{\symSafetyBBX}{\symBoundingVol^{safe}}
\DefineMathCmd{\symCollBox}{\symBoundingVol^{safe}}
\DefineMathCmd{\symSafetyRadius}{d_{safe}}

\DefineMathCmd{\symSensorRange}{d_{max}^{sense}}
\DefineMathCmd{\symFov}{\alpha_{s}}
\DefineMathCmd{\symHFov}{\alpha_{h}}
\DefineMathCmd{\symVFov}{\alpha_{v}}
\DefineMathCmd{\symSensorFov}{\symFov}
\DefineMathCmd{\symSensorRes}{R_{s}}
\DefineMathCmd{\symSensorHRes}{R_{sx}}
\DefineMathCmd{\symSensorVRes}{R_{sy}}
\DefineMathCmd{\symSensorRate}{rate_{s}}
\DefineMathCmd{\sensorPointcloud}{Z}
\DefineMathCmd{\cloudPoint}{\mathbf{z}}
\DefineMathCmd{\symProjVol}{V_p}
\DefineMathCmd{\symFovFn}{FoV}
\DefineMathCmd{\symVisFn}{Vis}
\DefineMathCmd{\symCoversFn}{\IndicatorFn^{covers}}
\DefineMathCmd{\symMap}{\mathcal{M}}
\DefineMathCmd{\symMapRes}{r_{\symMap}}
\DefineMathCmd{\symVoxel}{\boldsymbol{m}}
\DefineMathCmd{\symVoxelSpace}{\symMap}
\DefineMathCmd{\symVoxelSet}{\symMap}
\DefineMathCmd{\symVoxelSetFree}{\symVoxelSet^{free}}
\DefineMathCmd{\symVoxelSetOcc}{\symVoxelSet^{occ}}
\DefineMathCmd{\symVoxelSetUnk}{\symVoxelSet^{unk}}
\DefineMathCmd{\symChangedVoxels}{\delta \symMap}
\DefineMathCmd{\symOccupancyState}{\mathcal{O}}
\DefineMathCmd{\symOccStateUnk}{\symOccupancyState^{unk}}
\DefineMathCmd{\symOccStateFree}{\symOccupancyState^{free}}
\DefineMathCmd{\symOccStateOcc}{\symOccupancyState^{occ}}
\assign{\occLabel}{\textit{occ}}
\assign{\freeLabel}{\textit{free}}
\assign{\unkLabel}{\textit{unknown}}

\DefineMathCmd{\symMapOccupied}{\symMap^{occ}}
\DefineMathCmd{\symMapFree}{\symMap^{free}}
\DefineMathCmd{\symMapUnknown}{\symMap^{unk}}
\DefineMathCmd{\symViewableMap}{\mathcal{M}_{\Theta}}
\DefineMathCmd{\symTraversableMap}{\mathcal{M}_{Q}}
\DefineMathCmd{\symResidualMap}{\mathcal{M}_{res}}
\DefineMathCmd{\symLocalDiffMap}{\Delta \symMap}
\DefineMathCmd{\symLocalDiffVolume}{\Delta \symBoundingVol}
\DefineMathCmd{\symLocalOdomPath}{\Delta \Xi}
\DefineMathCmd{\symFrontier}{f}
\DefineMathCmd{\symFrontierSet}{\mathcal{F}}
\DefineMathCmd{\symSurfFrontierSet}{\symFrontierSet^{\symSurfSpace}}
\DefineMathCmd{\symVoidFrontierSet}{\symFrontierSet^{\symCSpace}}
\DefineMathCmd{\symFrontierSetRejected}{\symFrontierSet^{rej}}
\DefineMathCmd{\symFrontierSetSuspended}{\symFrontierSet^{rej}}
\DefineMathCmd{\symFrontierSetCovered}{\symFrontierSet^{cvr}}
\DefineMathCmd{\symFrontierSetNoncovered}{\symFrontierSet^{\overline{cvr}}}
\DefineMathCmd{\symView}{\lambda}
\DefineMathCmd{\symViewSet}{\Lambda}
\DefineMathCmd{\symViewpointNBV}{\symView_{nbv}}
\DefineMathCmd{\symViewSetAdded}{\Delta \symViewSet^{-}}
\DefineMathCmd{\symViewSetErased}{\Delta \symViewSet^{+}}

\DefineMathCmd{\symAgentLabel}{\symNodeLabel_{agent}}
\DefineMathCmd{\symHomeLabel}{\symNodeLabel_{home}}
\DefineMathCmd{\symKfLabel}{\symNodeLabel_{kf}}
\DefineMathCmd{\symTraversalLabel}{\symNodeLabel_{traversal}}
\DefineMathCmd{\symNBVLabel}{\symNodeLabel_{nbv}}

\DefineMathCmd{\symAgentVP}{\symView^{agent}}
\DefineMathCmd{\symHomeVP}{\symView^{home}}
\DefineMathCmd{\symCoverageVP}{\symView^{nbv}}

\DefineMathCmd{\symAgentViewset}{\symViewSet^{agent}}
\DefineMathCmd{\symHomeViewset}{\symViewSet^{home}}

\DefineMathCmd{\symViewsetKF}{\symViewSet^{kf}}
\DefineMathCmd{\symKfVP}{\symView^{kf}}

\DefineMathCmd{\symViewsetNBV}{\symViewSet^{nbv}}
\DefineMathCmd{\symNBVVP}{ \symView^{nbv}}

\DefineMathCmd{\symTraversalViewset}{\symViewSet^{\symCSpace}}
\DefineMathCmd{\symViewsetTraversal}{\symViewSet^{\symCSpace}}
\DefineMathCmd{\symTraversalVP}{\symView^{\symCSpace}}

\DefineMathCmd{\symViewSetOpen}{\symViewSet^{open}}
\DefineMathCmd{\symViewSetClosed}{\symViewSet^{closed}}

\DefineMathCmd{\symPoseSample}{ \bar{\symPose} }
\DefineMathCmd{\symPoseSampleSet}{ \bar{Q} }
\DefineMathCmd{\symViewPoseSample}{ \symPoseSample }
\DefineMathCmd{\symViewPoseSampleSet}{ \symPoseSampleSet }
\DefineMathCmd{\symViewSample}{ \bar{\symView} }
\DefineMathCmd{\symViewSampleSet}{ \bar{\symViewSet} }

\DefineMathCmd{\symGoalSpace}{ \symViewSet^{G} }
\DefineMathCmd{\symGoalUtility}{ \gamma }

\DefineMathCmd{\symAPN}{ \mathcal{G} }
\DefineMathCmd{\symGraph }{ \mathcal{G} }

\DefineMathCmd{\symApnVertexSet }{ \mathcal{V} }
\DefineMathCmd{\symApnVertex }{ v }
\DefineMathCmd{\symVertexView }{ \symView }
\DefineMathCmd{\symVertexPose }{ \symPose }
\DefineMathCmd{\symVertexID }{ uid }
\DefineMathCmd{\symVisitIndicator }{ \mathbb{1}^{open} }
\DefineMathCmd{\symVertexGain }{ \gamma }

\DefineMathCmd{\symHomeVertex }{ \symApnVertex^{home} }
\DefineMathCmd{\symAgentVertex }{ \symApnVertex^{agent} }
\DefineMathCmd{\symKfVertex }{ \symApnVertex^{kf} }
\DefineMathCmd{\symNbvVertex }{ \symApnVertex^{nbv} }
\DefineMathCmd{\symTraversalVertex }{ \symApnVertex^{\symCSpace} }

\DefineMathCmd{\symKfVertexSet }{ \symApnVertexSet^{kf} }
\DefineMathCmd{\symNbvVertexSet }{ \symApnVertexSet^{nbv} }
\DefineMathCmd{\symTraversalVertexSet }{ \symApnVertexSet^{\symCSpace} }
\DefineMathCmd{\symVertexSetOpen}{ \symApnVertexSet^{open} }
\DefineMathCmd{\symVertexSetClosed}{ \symApnVertexSet^{closed} }

\DefineMathCmd{\symNodeSet }{ \symApnVertexSet }
\DefineMathCmd{\symNode }{ \symApnVertex }

\DefineMathCmd{\symUID}{uid}
\DefineMathCmd{\symNodeLabel }{ \boldsymbol{l} }
\DefineMathCmd{\symNodeID }{ uid }

\DefineMathCmd{\symHomeNode }{ \symHomeVertex }
\DefineMathCmd{\symAgentNode }{ \symAgentVertex }
\DefineMathCmd{\symNodeGain }{ \gamma }

\DefineMathCmd{\symKfNode }{ \symKfVertex }
\DefineMathCmd{\symKfNodeset }{ \symKfVertexSet }

\DefineMathCmd{\symTraversalNode }{ \symTraversalVertex }
\DefineMathCmd{\symTraversalNodeset }{ \symTraversalVertexSet }

\DefineMathCmd{\symEdgeSet}{\mathcal{E}}
\DefineMathCmd{\symEdge }{e}
\DefineMathCmd{\symCollEdgeSet}{\symAdjEdgeSet^{\symCollSpace}}
\DefineMathCmd{\symCollEdge}{e^{\symCollSpace}}
\DefineMathCmd{\symEdgeCostFn }{c}
\DefineMathCmd{\symEdgeLabel }{ \boldsymbol{l}^{adj} }

\DefineMathCmd{\symEdgeBoundingVol}{OBB}
\DefineMathCmd{\symEdgeCollisionState}{\symCollisionState}
\DefineMathCmd{\symEdgeObsPoint }{\boldsymbol{p}^{obs}}

\DefineMathCmd{\symEdgeUpdateProbThresh }{ p^{\symEdge}_{update} }
\DefineMathCmd{\symEdgeSearchDist }{d^{\symEdge}_{max}}

\DefineMathCmd{\symViewVisMap}{ \Gamma }
\DefineMathCmd{\symFrontierVisMap}{ \Upsilon }
\DefineMathCmd{\symVisMap}{ \Gamma }

\DefineMathCmd{\symIndividualGain}{\mathcal{K}}
\DefineMathCmd{\symJointGain}{\mathcal{J}}
\DefineMathCmd{\symExclusiveGain}{\mathcal{I}}


\DefineMathCmd{\symHypergraph}{ \mathcal{H} }
\DefineMathCmd{\symHyperEdgeset}{ \mathcal{C} }
\DefineMathCmd{\symHyperEdge}{ \mathcal{H} }
\DefineMathCmd{\symHyperEdgeSubgraph}{ \mathcal{A} }
\DefineMathCmd{\symHyperEdgeNodeSet}{ \symNodeSet^{\symHyperEdgeset} }

\DefineMathCmd{\symHyperNodeset}{ \mathcal{C}^{\symHypergraph} }
\DefineMathCmd{\symHyperNode}{ \boldsymbol{c} }

\DefineMathCmd{\symViewCluster}{ \symHyperEdge }
\DefineMathCmd{\symViewClusterSet}{ \symHyperEdgeset }

\DefineMathCmd{\symClusterCostMatrix}{ \boldsymbol{m}^{\symHypergraph} }
\DefineMathCmd{\symViewpointCostMatrix}{ \boldsymbol{m}^{\symViewSet} }

\DefineMathCmd{\clusteringMinPts}{\rho_{c}}
\DefineMathCmd{\clusteringEps}{D_{c}}


\DefineMathCmd{\symSamplingMaxTime}{t^{sample}_{nbv}}
\DefineMathCmd{\symSamplingMaxAttempts}{ N^{attempt}_{nbv}}
\DefineMathCmd{\symSamplingMinPeriod}{dt^{sample}_{min}}
\DefineMathCmd{\symSamplingFactorLocal}{p^{\symView}_{local}}
\DefineMathCmd{\symSamplingFactorGlobal}{p^{\symView}_{global}}
\DefineMathCmd{\symFrontierSampleThresh}{\symSamplingMaxAttempts}

\DefineMathCmd{\symMaxTraversalSamples}{N^{sample}_{traversal}}
\DefineMathCmd{\symMaxTraversalSampleAttempts}{N^{attempt}_{traversal}}
\DefineMathCmd{\symMinTraversalSepRadius}{d^{sample}_{traversal}}

\DefineMathCmd{\symReachabilityUpdateVol}{\widehat{\symBoundingVol}}

\DefineMathCmd{\fnAddNode}{\textit{addNode}}
\DefineMathCmd{\fnAddEdge}{\textit{addEdge}}
\DefineMathCmd{\fnAddCollisionEdge}{\textit{cacheCollisionEdge}}
\DefineMathCmd{\fnAddUnknownEdge}{\textit{cacheUncertainEdge}}


\DefineMathCmd{\symOrderedPath}{\Pi}
\DefineMathCmd{\symViewClusterPath}{ \symOrderedPath^{\symViewCluster} }
\DefineMathCmd{\symViewPath}{ \symOrderedPath^{\symView} }
\DefineMathCmd{\symGlobalViewPath}{ \symViewPath }


\newcommand{\symRunTime}{\overline{T}}
\newcommand{\symRunTimeNinetyFive}{\overline{T}^{\symCoverageRatio^{95\%} }}
\newcommand{\symCycleTime}{\bar{t}^{cycle}}
\newcommand{\symCycleTimeAPN}{ \symCycleTime_{DFR} }
\newcommand{\symCycleTimePlanning}{ \symCycleTime_{plan} }
\newcommand{\symCycleTimeNav}{ \symCycleTime_{nav} }
\newcommand{\symExploreDist}{d_{expl}}

\newcommand{\symCycleTimeSampling}{\symCycleTime^{expand}}
\newcommand{\symCycleTimeRefine}{\symCycleTime^{refine}}
\newcommand{\symCycleTimeReach}{\symCycleTime^{reach}}
\newcommand{\symCycleTimeCluster}{\symCycleTime^{cluster}}

\DefineMathCmd{\symCoverageRatio }{\vartheta_{\symMap}  }
\DefineMathCmd{\symCoverageError }{\symCoverageRatio^{\varepsilon}}
\DefineMathCmd{\symVolumetricRate}{\ensuremath{\eta_{\symMap}}}
\DefineMathCmd{\symFreespaceExploreRate}{\ensuremath{\eta_{\symVoxelSetFree}}}
\DefineMathCmd{\symSurfaceExploreRate}{\ensuremath{\eta_{\symVoxelSetOcc}}}
\DefineMathCmd{\symVoxelSetOccFullCoverage}{\widehat{\symVoxelSet}^{occ}}
\DefineMathCmd{\symNodeDensityFactor }{ \vartheta_{\symNodeSet} }
\DefineMathCmd{\symEdgeDensityFactor }{ \vartheta_{\symEdgeSet} }

\newcommand{\symLocalChangeVolumeRadius }{r_{local}}




\section{Introduction} \label{introduction}

The general problem of online exploration and visual surface coverage of \textit{a priori} unknown structure or environment can be referred to as \textit{online sensor-based coverage planning} (OSCP). For this, a robot such as a Micro Aerial Vehicle (MAV) must efficiently discover the spatial and geometric structure of an initially unknown environment using an onboard depth sensor. The robot must traverse the environment to perceive the unknown space from different perspectives, accumulating the acquired sensor knowledge in a spatial map. OSCP is a prerequisite problem for a wide range of applications involving operation in an unknown environment, such as structural modeling and inspection, surveying, search and rescue, and many others \cite{quattrini2020exploration}. 

The \textit{global coverage problem} of OSCP is to achieve maximum coverage of the target surfaces as efficiently as possible. Naturally, this cannot be solved directly due to the lack of a priori knowledge, and can only be solved online in an incremental fashion. This leads to the \textit{incremental exploration problem} which represents an iterative action selection problem: given the current incomplete knowledge, determine the optimal action to increase the current knowledge. As each action is executed, the incremental objective is recursively solved using feedback from the added environment knowledge until the global exploration objective has been achieved. 

\section{Related Work} \label{s:relatedWork}

The purpose of online exploration and coverage can vary between different applications and tasks. For example, some applications seek knowledge of the traversable space within an unknown environment for subsequent navigation tasks, while others may seek detailed coverage of the surfaces for 3D modelling or inspection purposes. It is important to recognize such differences in the intended application, as this can greatly influence how the problem is approached and how its performance is evaluated.

\smallskip\noindent 
\textbf{Frontier-based exploration}: \noindent 
Autonomous exploration was pioneered by \cite{Yamauchi1997} by introducing the now well-known concept of spatial frontiers. Frontiers represent boundaries within a partially built map between unknown space the robot seeks to observe, and the free space the robot can use to make the observations. The original frontier exploration algorithm, referred to as classical frontier exploration, was implemented for a mobile ground robot building a 2D occupancy grid map. The algorithm selects the closest frontier as the goal, and navigates towards the goal while using reactive collision-avoidance. Upon arrival, a new sensor scan is acquired of the region and added to the map, repeating the process until no unvisited and reachable frontiers remain.

Many extensions have been proposed to the initial frontier-based approach, including more efficient frontier detection methods \cite{keidar2014efficient, topiwala2018frontier} and extensions for 3D maps \cite{zhu20153d}. However, a significant drawback of classical frontier exploration is that frontiers indicate only the \textit{existence} of adjacent unknown space, but not the quantity or quality. Using a frontier location directly as the navigation goal ignores sensor's measurement range, thus causing inefficient and wasteful motions. Furthermore, a frontier location near surfaces generally do not represent a feasible goal for a robot due to collision with the surface obstacle, making their direct use in this way ineffective for surface coverage tasks. 

\smallskip\noindent 
\textbf{Next-best-view (NBV) sampling}: \noindent 
Exploration can be effectively modeled as an extension of the Next-Best-View (NBV) problem introduced by \cite{connolly1985determination}, which can overcome several of the drawbacks associated with classical frontier-based approaches. Here, a \textit{view} refers to a hypothetical pose of the sensor apparatus used to predict and analyze the spatial information expected to be visible if the real sensor were to be placed at this pose. The expected visible information is then said to be \textit{covered} by the view. 

The classical NBV problem assumed full prior knowledge of the target object is given to facilitate the search and evaluation of NBVs, where the objective was to find a minimum set of views that maximizes coverage of the known surfaces of the object model. This premise can be adapted for online exploration tasks by instead evaluating views according to currently unknown parts of the environment model, rather than the known parts.

NBV-based exploration methods typically utilize a generate-and-test paradigm which apply sampling techniques necessary to discretize the continuous configuration space into a finite set of candidate views for analysis \cite{scott2003view}. The quality of a view is evaluated according to some measure of its \textit{information gain} (IG), which quantifies the new spatial information potentially observable from the view \cite{stachniss2005information, heng2015efficient, kaufman2016autonomous, song2018surface}. A cost metric is additionally used to evaluate the expected effort for the robot to visit the view (e.g. time or energy). Most critical differences between existing NBV approaches occur within the sampling strategy for generating view candidates, and the formulation of metrics for analyzing and comparing candidates for goal selection. 

Information gain is commonly computed volumetrically by finding the expected amount of unknown space visible from a view \cite{okada2015exploration, dang2019graph, kompis2021informed}. This necessarily involves checking for occlusions within the known space using techniques like raycasting, which incurs high computational complexity that can rapidly increase with various factors like map resolution, sensor field of view, and sensing range. This limits the number of distinct views that can be practically evaluated within a given time period. The high complexity also make it difficult to analyze overlapping or mutual information between views, such that most approaches treat the gain as an independent value that prevents an understanding of the unique gain contributions of each view within a group.


\smallskip\noindent 
\textbf{Tree-based planning}: \noindent 
Tree-based methods organize sampled views as vertices in a geometric tree where directed edges between vertices represent feasible paths between views. The RH-NBVP approach of \cite{bircher2016receding, bircher2018receding} applies rapidly-exploring random tree (RRT) to grow a tree rooted at the robots current position. Each node in the tree is weighted according to their predicted information gain based on how much unknown space lies within the view. Cost weights are aggregated along each branch, and the leaf node with the highest value is used to identify the best branch to explore, iteratively repeating the process in a receding horizon fashion. This has become a well-known approach and is often used as a baseline for comparative analysis \cite{selin2019efficient, dai2020fast, cieslewski2017rapid}.

A hybrid approach that combines both frontier-based and NBV-based techniques was introduced by \cite{selin2019efficient}, referred to as AEP. It combines the RH-NBVP strategy for local planning, while switching to frontier-based planning for global search when local planning fails to find informative views. FFI \cite{dai2020fast} is also a hybrid approach that uses an efficient frontier clustering strategy to guide view sampling.

A significant drawback of tree-based planning is the difficulty in preserving the previously computed tree structure as the robot navigates to each goal. The RH-NBVP approach builds a new tree each iteration, discarding the previously built structure that may still contain useful knowledge. Other approaches attempt to transfer as much of the previous tree structure as possible by rewiring its edges to initialize the construction of a new tree. Since tree-based methods are rooted at the robots position, they tend to become increasingly inefficient over larger distances, making it difficult to handle dead-end or backtracking cases.

\smallskip\noindent 
\textbf{Graph-based planning}: \noindent 
Various approaches have utilized graph structures that can overcome some of the limitations and drawbacks of trees. The approach of \cite{witting2018history} builds a history graph that stores previously visited positions and their edge connections. These are used as potential seed points for RRT, which allows a tree to be grown from different positions across the map, rather then just from the robot position. An approach using Rapidly-Exploring Random Graphs (RRG) was presented in \cite{dang2019graph} for exploration of subterranean environments. A Probabilistic Roadmap (PRM) strategy was used by \cite{xu2021autonomous} to build a graph of feasible configurations and paths over the map as it is explored.  

\smallskip\noindent 
\textbf{Topological maps}: \noindent 
Topological maps have been applied by recent works which aim to reduce the planning complexity through the compact representation provided by a topological map. Topological maps can be considered as an extension to graph-based methods, where vertices represent some volumetric sub-map, or \textit{place}, and edges represent the adjacency or reachability between places. This coarse and abstracted representation is more efficient for handling large-scale environments, which can become intractable to explore online using alternative approaches. However, they usually lack sufficient metric knowledge for direct use in navigation.

\cite{silver2006topological} used a topological map for exploration of underground mines using a ground robot. The regions of intersection between passageways were represented as nodes, and exploration was planned along the edges between nodes. A more recent approach proposed by \cite{yang2021graph} also uses a topological map for subterranean exploration. Convex polyhedrons are used to estimate distinctive exploration regions (DER-s) which are added as graph nodes to the map. Each DER represents an enclosed 3D volume of the map like an enclosed room or corridor, providing the planner with knowledge of high-level intent such as moving between distinctive rooms or regions. Other approaches have applied segmentation algorithms to identify the separation of distinct exploration regions like rooms of a building \cite{fermin2017incremental}. 

\smallskip\noindent 
\textbf{Myopic greedy planning}: \noindent 
The majority of existing methods compute navigation goals using myopic planning strategies that greedily optimize the cost of the next single planning decision \cite{palazzolo2018effective, dai2020fast}, or within a limited planning horizon \cite{bircher2016receding, selin2019efficient}. Some works allow planning over the full map, but still use greedy search for the decision making. These are sometimes referred to as global planning methods, but we clarify they are still considered myopic.

Myopic strategies bias exploration toward regions with high information gain, while ignoring small gains even if they are closer. This bias can frequently create regions of incomplete coverage when a high gain goal leads the exploration away from the current region before it is fully mapped. This can also result in frequent back-and-forth oscillation between goals, or require re-visitation of these regions after the robot has traveled a significant distance, backtracking over potentially large distance. This greatly reduces efficiency, and can result in sparse coverage gaps or failure to fully explore an environment within an allowed time limit, especially over large-scales.

A relatively small number of works have recently attempted to overcome the drawbacks of greedy planning using non-myopic planning strategies. This has been formulated using the Traveling Salesman Problem (TSP) \cite{meng2017intelligent, song2020online}, but often relies on prior map knowledge \cite{bircher2016three, shang2020co}. 

A sector decomposition approach was presented by \cite{song2020online}, which partitions the map into a set of convex sectors used to compute a TSP sequence. However, the sector decomposition method can be computationally expensive, especially for finer map resolutions, which can greatly decrease the update rate of the map and planning. Additionally, sectors form an exact partitioning of the space, which can make the geometric properties of the resulting sectors difficult to control, and may not effectively handle large-scale and complex environments.

\smallskip\noindent 
\textbf{Environment and task-specific approaches}:  \noindent 
Simplifying or restrictive assumptions are sometimes made on the operational environment. This can include indoor operation, or reliance on certain regular geometric features, e.g. room structures used for segmentation. Some applications are intended to operate in relatively obstacle-free environments, such as outdoors or underwater \cite{ellefsen2017multiobjective}, which contain an abundance of free-space that greatly simplifies collision checking and other sub-tasks. Assumptions can significantly restrict the practicality of many approaches for general use, or require fine tuning of parameters between different environments to achieve their rated performance.

\parName{\textbf{Limitations of existing approaches:}}
Limitations of existing approaches are summarized as follows:
\begin{itemize}
    \item greedy and myopic planning strategies that focus on the incremental exploration objective, but fail to consider the global one,
    \item non-generalized approaches that are limited to small-scale environments, or specialized for specific environments or conditions (e.g. subterranean or building-like structures),
    \item most approaches succumb to high computational costs: 
    \begin{itemize}
        \item they do not scale well with respect to environment size or map resolution,
        \item the ability to quickly replan on added knowledge diminishes, where a suboptimal plan is fully executed before replanning,
        \item reduced velocities are often required to compensate for low planning rates,
        \item frequent stop-and-go motions can occur.
    \end{itemize}
\end{itemize}
We observe that there is not sufficient attention to the underlying data management issues of the general OSCP problem in the robotics research community, which could be due to limited research funding and development cycle and that data infrastructure was not a focus. There is also a lack of open-source software to help reducing efforts that researchers have to put into developing a good data management system. The aforementioned limitations of existing approaches are the results of that. However, for many realistically large-scale OSCP tasks, it is critical to have smart and sophisticated data management system, requiring careful conceptual, algorithmic, and data structure design and efficient software engineering solutions. 

\section{Contributions} \label{s:Contributions}

This work is motivated to alleviate some of the limitations of the existing approaches in handling the OSCP problem. We focus first on how to dynamically compute and maintain the accurate global knowledge necessary to a non-myopic planning algorithm, since this represents a significant bottleneck in terms of computational complexity and exploration quality in the existing work. Our key contributions are as follows:

\begin{itemize}
\item A novel dynamic multi-layer topological graph designated as the \textit{Active Perception Network} (APN). The APN serves as a global hierarchical roadmap over the spatial map that accumulates the incrementally computed knowledge of the exploration state space. is defined and organized around adaptive nodes to best represent the perceptual and actionable environment knowledge discovered to minimize the complexity, which allows it to be efficiently accessed and searched for planning purposes.

\item A dynamic update procedure referred to as \textit{Differential Regulation} (DFR) to incrementally build and refine the APN as environment knowledge is increased. This procedure addresses the complexity of updating the APN as its size and the map scale increase, while ensuring sufficient global knowledge is maintained for effective planning.

\item A non-myopic planning approach denoted as APN-Planner (APN-P) that demonstrates how the APN can be leveraged to compute and adaptively refine a globally informed exploration sequence.
\item A detailed performance analysis and comparison to existing approaches among the state-of-the-art.
\item An open-sourced release of the APN, DFR, and planning implementations, and a programming framework for the development of generalized autonomous exploration approaches that was developed and used for all our implementations.
\end{itemize}

\section{Problem Formulation} \label{sec:problemForm}
We assume exploration is performed using an MAV equipped with an onboard depth sensor (e.g. stereo-visual, RGB-D, or LiDAR) to perceive 3D space, noting that other systems such as mobile ground robots could also be utilized without loss of generality. We define the following terms and symbols to facilitate  the description of our approach:

\parName{\textbf{Environment and map model:}}%
Let $\symWSpace \subset \mathbb{R}^{3}$ represent the bounded 3D space of the operational environment, referred to as the \textit{world}. The solid structures and objects of the world represent \textit{occupied} space $\symWSpaceOcc \subset \symWSpace$, while the remaining volume is defined as \textit{free-space} $\symWSpaceFree \subset \symWSpace$, such that $\symWSpace \equiv \symWSpaceFree \cup \symWSpaceOcc$. 

The intersection boundaries between occupied and free-space define the surface manifolds, $\symSurfSpace \subset \mathbb{R}^{2}$. Surface manifolds are assumed to be visually opaque, and a surface point is considered optically visible from a point $\symPnt \in \symWSpaceFree$ only if no occupied space lies between the surface and $\symPnt$. Otherwise, the surface is considered to be occluded from $\symPnt$. 

A spatial map $\symMap$ is used to store the environment state knowledge as it is discovered from sensing. We assume the use of a 3D grid-based occupancy map $\symMap = \{ \symVoxel_0, \ldots, \symVoxel_m \}$, though other map models could also be used without loss of generality (e.g. Signed Distance Field (SDF) \cite{oleynikova2017voxblox}). $\symMap$ partitions $\symWSpace$ by a set of non-overlapping cubic volumes $\symVoxel \in \mathbb{R}^{3}$, known as voxels. The minimum edge length of a voxel dictates the map resolution, $\symMapRes$.

Each voxel stores the occupancy probability of its volume, which is updated from sensor measurements depending on whether occupied or free-space was observed. The probability value is discretized by an occupancy state $\symOccupancyState \in \{ \symOccStateUnk, \symOccStateOcc, \symOccStateFree \}$, where $\symOccStateUnk$ indicates the state is \textit{unknown}. As sensor measurements are integrated, the state is classified as either $\symOccStateOcc$ or $\symOccStateFree$ to indicate, respectively, whether the voxel belongs to the set of occupied voxels $\symVoxelSetOcc \subseteq \symMap$, or free-space voxels $\symVoxelSetFree \subseteq \symMap$. The set of occupied voxels are given as $\symVoxelSetOcc \subseteq \symMap$, the set of free voxels is given by $\symVoxelSetFree \subseteq \symMap$, and the set of unknown voxels is given by $\symVoxelSetUnk \subseteq \symMap$, with the initial map state given as $\symMap \overset{\mboxTiny{init}}{=} \symVoxelSetUnk$.

Spatial frontiers, $\symFrontierSet$, are detected from $\symMap$ by identifying unknown voxels with an adjacent free voxel. Frontiers that are also adjacent to an occupied surface voxel are further classified as \textit{surface frontiers}, $\symSurfFrontierSet$. Those that are adjacent only to free space are classified as \textit{void frontiers}, $\symVoidFrontierSet$, such that $\symFrontierSet \equiv \symSurfFrontierSet \cup \symVoidFrontierSet$. These distinctions are made according to the goal of achieving complete surface coverage, where surface frontiers help to identify where surface coverage is incomplete. 

\parName{\textbf{Robot model:}}
The robot agent is modeled by a rigid body with pose configuration $\symRobotPose(t) = ( \symPos, \symEulerVec ), \ \symPose \in SE(3)$ at time $t$, where $\symPos \in \symReals^{3}$ is the position vector and $\symEulerVec = \{ \symRoll, \symPitch, \symYaw \}$ is the orientation vector represented by roll, pitch, and yaw Euler angles, respectively. Additional parameters $\symMaxVel$ and $\symMaxYawRate$ are used to specify the maximum allowable velocity and yaw rate, respectively. A spherical volume $\symSafetyVol$ centered at $\symPos$ with radius $\symSafetyRadius$ is defined, where $\symSafetyRadius$ specifies the minimum obstacle separation distance for safe operation. 

\parName{\textbf{Sensor model:}}
The robot's depth sensor is modeled by the parameter vector $[ \symSensorRes, \symSensorFov, \symSensorRange ]$. $\symSensorFov = [ \symHFov, \symVFov ] \in (0, 2 \pi ]$ is the maximum angular field of view (FoV) on the horizontal and vertical dimensions of the sensor, and $\symSensorRes = [ \symSensorHRes, \symSensorVRes ]$ is the maximum spatial resolution. $\symSensorRange \in \mathbb{R} $ is the maximum effective sensing range that surface points can be accurately detected by the sensor. This value corresponds to the physical limitations of the sensor, where distances greater than $\symSensorRange$ either cannot be measured, or are rejected due to loss of accuracy.

The sensor parameters can be combined with a pose $\symPose$ to form a projection model $\symView \in \symViewSet$, referred to as a \textit{viewpose}. The projected space from $\symView$ is described by the subset of rays that pass through the view's origin $\symPos$, constrained by the intervals $[ \symPitch \pm \symVFov / 2 ]$ and $[ \symYaw \pm \symHFov / 2 ]$ of the unit-sphere. The length of each ray is constrained by $\symSensorRange$. The projected space defines the view volume of a viewpose, and a location within the view volume is considered visible if there are no occlusions between it and the origin. This provides the basis for making visibility queries and predictions on the expected information gain.

\parName{\textbf{Reachable configuration space:}}
Given the robot's initial position $\symRobotPosn_{0}$, the \textit{reachable configuration space} $\symCSpace \subset \symReals^{3}$ is a metric space defined by all admissible configurations path-connected to $\symRobotPosn_{0}$. As a precondition, a configuration is considered \textit{admissible} if it does not intersect any occupied space within distance $\symSafetyRadius$. It is then considered \textit{reachable} if there exists a simply-connected path of admissible configurations from $\symRobotPosn_{0}$. The distance between two reachable points is quantified by a metric value $\symReachCost \in \symReals$.

\parName{\textbf{Goal space:}}
The surfaces that can possibly be covered at any point during exploration is inherently restricted to a subset $\symVisibleSurfs \subseteq \symSurfSpace$ which are visible from some viewpose $\symView$ constrained by $\symCSpace$. The \textit{goal space} $\symGoalSpace \subset \symViewSet$ is then defined as the set of feasible configurations that contribute some amount of coverage of $\symVisibleSurfs$, quantified by a gain  metric, $\symGoalUtility \in \symReals$.

\parName{\textbf{Exploration state space:}}
The \textit{exploration state space}, $\symXStateSpace$, refers to the collectively available knowledge necessary to solve the incremental exploration problem. This mainly consists of the robot pose $\symRobotPose$, spatial map $\symMap$, and $\symFrontierSet$, which are considered as independent time-varying input variables. It additionally includes the reachable C-Space, $\symCSpace$, and goal space, $\symGoalSpace$, which are dependent variables computed from the input data.

\parName{\textbf{Myopicity:}}
A planning strategy operates on the exploration state space to search for the optimal goal $\symPose^{g} \in \symGoalSpace$ for navigation, where the myopicity corresponds to the length of its planning horizon. A myopic strategy typically uses greedy search techniques which treats each goal or action as independent of the others, greedily selecting the best one. They may also constrain the search to only some local sub-region of the map, rather than considering its full extent. Myopic strategies focus on the optimization related to the incremental exploration problem, which are not necessarily optimal with respect to the global coverage objective.

In contrast, a non-myopic strategy searches over a long horizon that spans most or all of the available map. It additionally considers how the particular selection of a goal and its associated action may alter the future exploration state space. This involves search and evaluation over ordered sequences of actions, rather than each action individually. This results in solutions that are more optimal with respect to the long-term global coverage objective.

\section{Approach Overview}\label{sec:approach}

In this work, we address how to build a reusable exploration state space $\symXStateSpace$ that is adaptively maintained over the full spatial map as it is built concurrently. The iteratively built exploration space is then used to facilitate efficient non-myopic planning. We seek an approach that generalizes well to different environments with varying complexities and geometric characteristics, and efficiently scales to large-sized environments that cannot be effectively solved by myopic approaches.

To achieve this goal, we introduce a novel graph-theoretic information structure named the \textit{Active Perception Network} (APN) to model the exploration state space data, detailed in Section \ref{sec:approach:apn}. A key feature of the APN is a hierarchical representation over its configurations that helps to reduce its size complexity and enables variable-resolution planning as the map increases in scale. Another focus of the APN is the storage and organization of the contained data, such that dynamic changes can be efficiently made to any of its contents as its size increases, while also maximizing the low-level efficiency for search and query operations. Some of these details are related to software, data structures, and other implementation challenges, which are beyond the scope of this work. Instead, the APN will primarily be discussed from a modeling perspective, with some additional implementation details provided in the appendix.

We additionally introduce the process of \textit{Differential Regulation} (DFR) in Section \ref{sec:approach:dfr}, which operates on the APN to modulate its state with respect to the increasing map knowledge. DFR consists of sampling-based methods for increasing knowledge of the goal space and reachable space. A novel approach for information gain analysis is utilized that enables the individual and mutual information gain of the APN to be efficiently computed, which is leveraged to accelerate informative view sampling, pruning, and refinement. 

DFR exploits the incremental nature of map building where each sequential map update induces changes that occur only within a relatively small local region of bounded volume, independent of the total map size. With this insight, these incremental changes are tracked and cached using difference-awareness and memoization strategies to greatly reduce the computational overhead necessary to update the APN. This allows more discrete updates to be performed in a given time period, increasing the completeness and accuracy of each update. The ability to quickly perform each update is also critical to ensure the size of the map changes remain small, since the complexity of each update scales with the size of the changes. 

An anytime exploration planner is presented in Section \ref{sec:approach:viewPlanning}, which demonstrates the use of the APN to efficiently compute non-myopic global exploration sequences. The hierarchical representation of the APN is leveraged to first compute a global topological exploration plan over the full map. The beginning of the global plan is then locally optimized at a higher-resolution. Similar to the difference-aware approach used by DFR, sequential changes to the APN typically occur within locally bounded regions which are leveraged to initialize new planning instances from previous results. This allows optimizations to achieve faster convergence despite the increasing size of the map and APN.

\begin{figure}[tbp]
\centering
\captionsetup{justification=centering}
\includegraphics[width=0.85\columnwidth, trim={0 0 0 0 },clip]{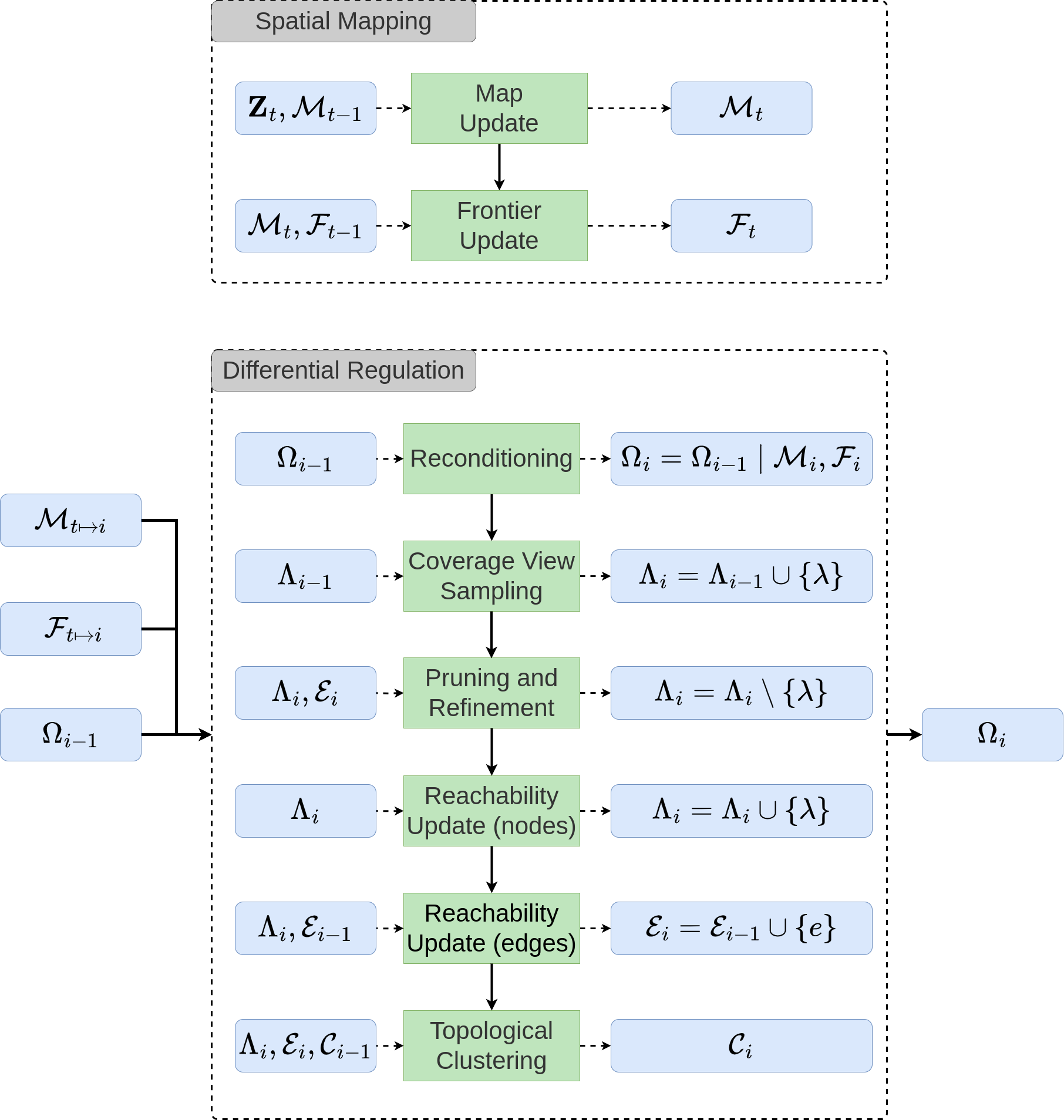}
\caption{Mapping, frontier detection, and Differential Regulation process pipelines used to update the APN.}
\label{f:DFR}
\end{figure}

The iterative exploration pipeline is illustrated in Fig. \ref{f:DFR}, which consists primarily of two asynchronous processing loops. The first loop is dedicated for spatial mapping to allow continuous integration of the sensor measurement data, $\mathbf{Z}_{t}$, at high frequency. Frontier detection is performed after each map update, which operates only on the state-changed voxels that resulted from the update. This minimizes the complexity required to maintain the global frontier set, and provides a constant upper complexity bound that remains independent of the total map size. The second loop concurrently performs DFR to update the APN, which then serves as the input for replanning the current exploration solution. Further details of each DFR subroutine will be provided in Section \ref{sec:approach:dfr}.

\section{Active Perception Network (APN)} \label{sec:approach:apn}

The Active Perception Network (APN) serves as a topological roadmap that stores the unified knowledge of the dynamically exploration state space. Its fundamental structure is represented by a hypergraph

\begin{equation} \label{eq:apnGraph}
\begin{gathered}
    \symGraph = (\symApnVertexSet, \symEdgeSet, \symHyperEdgeset),
\end{gathered}\end{equation}

\noindent
where $\symApnVertexSet = \{ \symApnVertex_i \}_{i=1,\dots,n}$ is the set of graph nodes and $\symEdgeSet = \{ \symEdge_{u,v} \}_{u,v \in [1,n]}$ is the set of traversal edges between nodes. The nodes have a bijective mapping to a codomain of viewposes, $\symApnVertexSet \hookrightarrow \symViewSet$, where the terms node and viewpose may also be referred to interchangeably. $\symApnVertexSet$ is decomposed by a set of hyperedges $\symHyperEdgeset = \{ \symHyperEdge \} \in \symPowerSet(\symApnVertexSet) $, where $\symPowerSet$ is the power set. Each hyperedge $\symHyperEdge \subseteq \symApnVertexSet$ contains a disjoint subset of $\symApnVertexSet$ as a multi-level hierarchy.

\begin{figure*}[!htbp]
\centering
\captionsetup{justification=centering}
	\subfloat[Visualization of the APN graph in terms of its node classifications (see figure legend) and traversal edges (blue lines). \label{f:approach:APNGraph}]{
		\centering
		\includegraphics[width=0.4\textwidth]{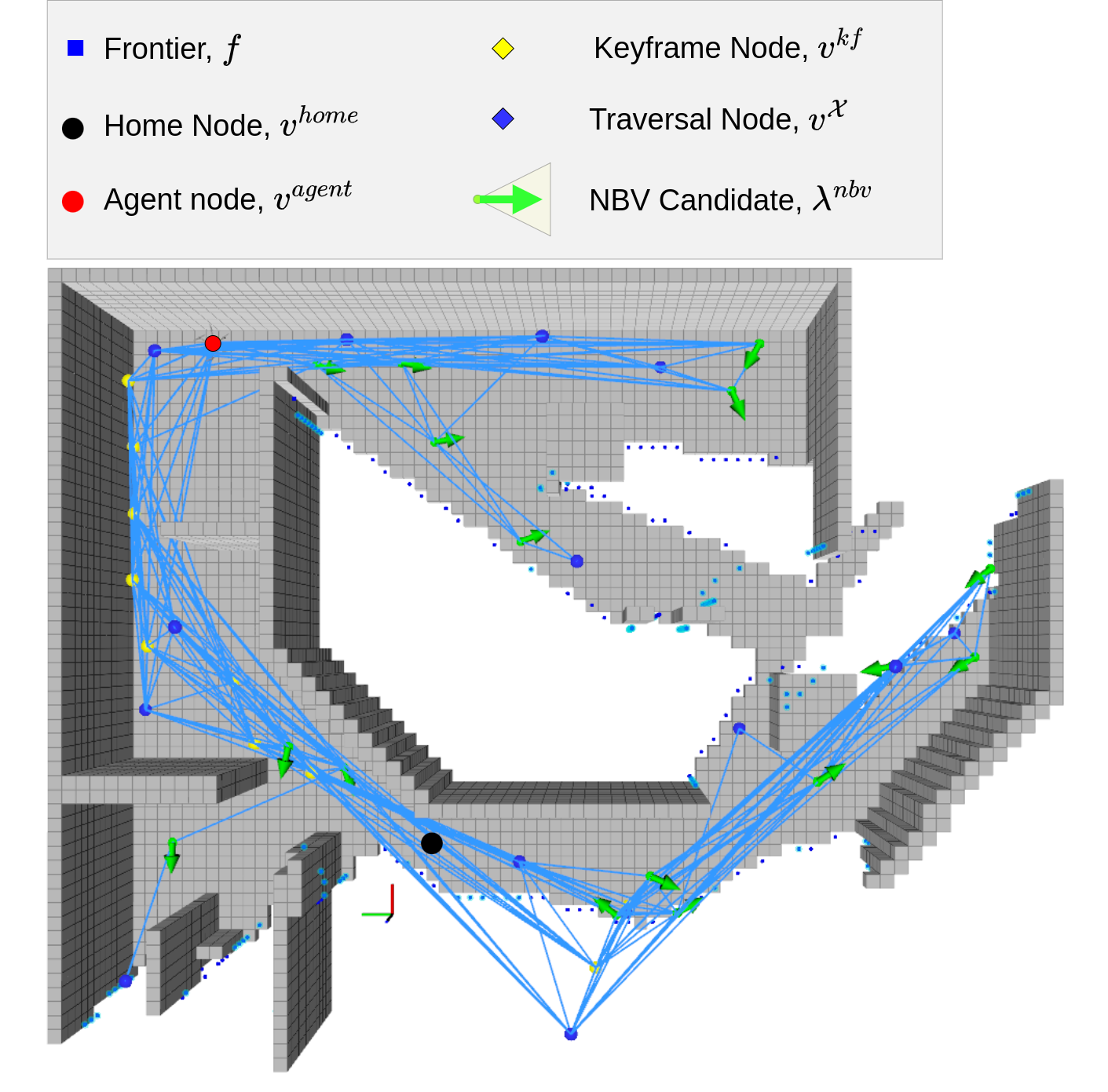}
	}
	\qquad
	\subfloat[APN hyperedge clusters, visualized by a bounding box enclosing the contained nodes. Each cluster forms an induced subgraph over the intra-cluster edges (orange lines). \label{f:approach:APNClusterGraph}]{
		\includegraphics[width=0.4\textwidth]{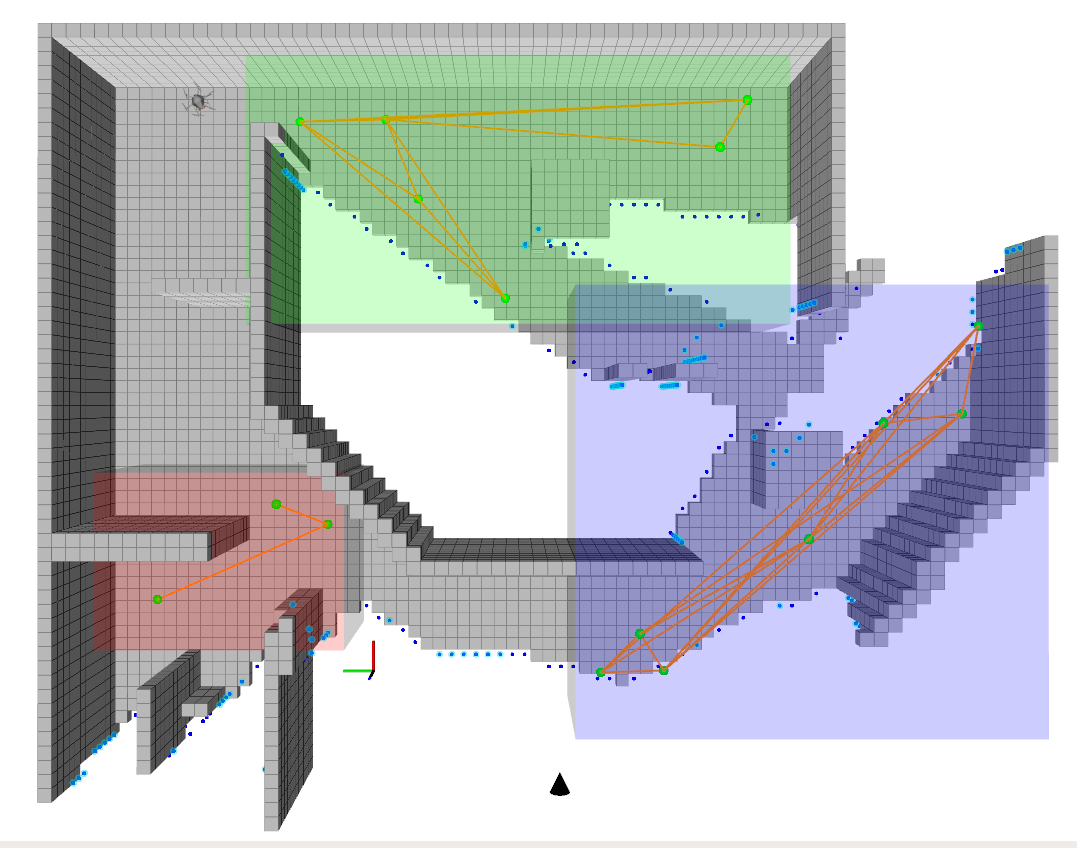}
	}
	\\
\caption{Depictions of the APN composition.}
\label{f:APN:diagrams}
\end{figure*}

\parName{\textbf{Graph nodes}, $\symApnVertexSet$:}
\noindent Each node $\symApnVertex_i \in \symApnVertexSet$ represents a viewpose information structure that consists of the tuple

\begin{equation} \begin{gathered} 
\symApnVertex_{i} = \{ \symVertexPose_{i}, \symVertexGain_{i}, \symVisitIndicator_{i} \},
\end{gathered}\end{equation}

\noindent
where $\symPose_{i} $ is its pose which has an associated viewpose $\symPose_{i} \mapsto \symView_{i}$, and $\gamma_{i} \in \symReals$ is a reward metric that quantifies the expected information gain available from $\symView_{i}$. The node's visitation state is stored by a Boolean indicator $\symVisitIndicator_{i} : \symApnVertex_{i} \mapsto \mathbb{B}$, corresponding to whether the robot has visited the pose of $\symApnVertex_{i}$. A \textit{true} value indicates the node is unvisited, also referred to as \textit{open}, and is otherwise referred to as \textit{closed} if it has already been visited. This is used to discriminate between the open set of nodes $\symVertexSetOpen$ which can represent goal candidates, and the closed set $\symVertexSetClosed$ of nodes which have already been visited. 

Several important classifications are defined over $\symApnVertexSet$ based on their properties. These provide an increased understanding of how the network can serve different tasks. These are summarized as follows:

\begin{itemize}
    
    \item A unique node $\symAgentVertex \in \symApnVertexSet$, referred to as the \textit{agent node}, is used to represent the robot and is dynamically updated with the robot pose as it changes over time. The robot's initial pose $\symRobotPose_{0}$ is used to define the \textit{home state}, represented by a unique node $\symHomeVertex$ that remains fixed over the lifetime of the APN. 
    
    \item The previously traversed path of the robot is represented by a path-connected set of \textit{keyframe nodes}, $\symRobotPose_{0:t} \mapsto \{ \symKfVertex_{0:k} \}  \in \symKfVertexSet $, rooted at the home state, $\symKfVertex_{0} = \symHomeVertex$. Keyframe nodes are added in intermediate intervals once the robot has traveled a minimum distance from the last keyframe.
    
    \item Unvisited nodes with positive information gain are classified as \textit{NBV candidate nodes}, represented by the set $\symNbvVertexSet = \{ \symApnVertex \in \symVertexSetOpen : \symVertexGain (\symApnVertex) > 0 \}$. A Next Best View (NBV) node represents a subgoal candidate for for navigation and planning that is expected to increase map knowledge.
    
    \item The remaining \textit{traversal nodes}, $\symTraversalVertexSet = \symApnVertexSet \setminus \symNbvVertexSet$, mainly serve to preserve the accumulated knowledge of the reachability space and its connectivity, but not expected to increase map knowledge.
    
\end{itemize}

\parName{\textbf{Graph edges}, $\symEdgeSet$:}
Each edge $\symEdge_{u,w} \in \symEdgeSet$ corresponds to the pair of nodes $\langle \symApnVertex_{u}, \symApnVertex_{w} \rangle$, and stores various analytical information of the traversal space between the pair as follows:

\begin{equation} \begin{gathered} \label{eq:edgeCost}
\symEdge_{u,w} = \{ d^{\symPos}, d^{\symYaw}, L, \symEdgeBoundingVol, \symEdgeCollisionState, \symEdgeObsPoint \},
\end{gathered}\end{equation}

\noindent
where $d^{\symPos}$ and $d^{\symYaw}$ are the Euclidean distance and the orientation angle distance, respectively, between $( \symApnVertex_{u}, \symApnVertex_{w} )$. $L$ is the evaluated cost metric value to traverse the edge given the maximum velocity $\symMaxVel$ and yaw rate $\symMaxYawRate$, defined by:

\begin{equation} \begin{gathered} 
L( \symEdge_{u,w} ) = max \left( \frac{d^{\symPos}(\symEdge_{u,v})}{\symMaxVel}, \frac{d^{\symYaw}(\symEdge_{u,v})}{\symMaxYawRate} \right).
\end{gathered}\end{equation}

Each edge also stores the Oriented Bounding Box (OBB) enclosing the endpoints, and the collision state of the space contained in the OBB is stored by $\symEdgeCollisionState : \symEdgeBoundingVol \rightarrow \{ free, unk, obs \}$. $\boldsymbol{p}^{obs}$ is used as a memory cache that stores any uncertain voxels found from previous collision checks. This allows for lazy evaluation during future checks by first checking if these discrete voxels have changed, rather than the full OBB volume, to greatly reduce complexity.

\parName{\textbf{Hyperedge clusters}:}
A set of hyperedges $\symHyperEdgeset \subset \symPowerSet(\symApnVertexSet) $ forms a topological decomposition of $\symGraph$, providing a representation with reduced size and complexity. A hyperedge $\symHyperEdge \in \symHyperEdgeset$ represents a cluster of nodes $\{ \symApnVertex \} \subseteq \symApnVertexSet$ grouped according a similarity measure between the nodes, such that $\symHyperEdgeset$ is a partitioning of $\symApnVertexSet$ into disjoint subsets $\{ \symHyperEdge \}$. Each hyperedge is modeled by the following:

\begin{equation} \begin{gathered} 
\symHyperEdge_{i} = \{ \symHyperEdgeNodeSet_{i}, \symHyperEdgeSubgraph_{i}, \symBoundingVol_{i}, \symPos_{i} \},
\end{gathered}\end{equation}

\noindent
where $\symHyperEdgeNodeSet_{i}$ is the set of nodes belonging to $\symHyperEdge_{i}$, with the centroid of the contained nodes given by $\symPos_{i}$ and its bounding volume given as $\symBoundingVol_{i}$. $\symHyperEdgeSubgraph_{i} = G[ \symHyperEdgeNodeSet_{i} ] $ is the vertex-induced subgraph formed by each cluster containing the clustered nodes $\symApnVertex \in \symHyperEdge$ and the induced edges $( \symEdge_{u,w} \in \symEdgeSet : \symApnVertex_{u}, \symApnVertex_{w} \in \symHyperEdgeNodeSet_{i})$ with both endpoints belonging to $\symHyperEdgeSubgraph_{i}$. 

The induced edges of a cluster subgraph $\symHyperEdgeSubgraph$ are referred to as its \textit{interior edges}, while the remaining edges that connect nodes between different clusters are referred to as \textit{exterior edges}. The efficiency of global search queries and traversal through $\symGraph$ can greatly increased by traversing between subraphs using their exterior edges, using the interior edges of the subgraphs to perform local operations as needed.

\section{Differential Regulation} \label{sec:approach:dfr}

The APN is incrementally built by the process of \textit{Differential Regulation} (DFR), which manages how information is added, removed, or modified in the APN with respect to the concurrently built spatial map. DFR evaluates the APN according to a set of objectives and constraints conditioned on the current map, and executes a set of modifying procedures on the APN as needed to ensure they remain satisfied as the map evolves. 

The broad purpose of the DFR procedures is to a) re-evaluate map-dependent analytical measures to ensure their accuracy (e.g. information gain of existing nodes), b) add node and edge elements to increase the completeness of the network while pruning redundant or overcomplete elements, and c) recompute the topological clustering of the updated graph state. A diagram of these procedures is shown in Fig. \ref{f:DFR}, and detailed in the following subsections.

\subsection{Reconditioning} \label{sec:approach:updateInit}

Each DFR cycle $i$ begins at a time $t$ with the latest spatial map $\symMap_{t(i)}$, frontiers $\symFrontierSet_{t(i)}$, and robot pose $\symRobotPose_{t(i)}$. The first task is to determine the local differences of these variables to their states from the previous cycle $t(i-1)$. Each incremental map update reports the set of state-changed voxels, which are accumulated in a local cache $\symLocalDiffMap$ with its bounding volume $\symLocalDiffVolume$. This is defined as the \textit{local difference neighborhood} and is used to inform various APN update procedures about where state-changes have occurred, described further in the next subsections.

Each regulation cycle then begins by updating the pose of the agent node $\symAgentVertex$ and its local edges. The length of the local path is then checked and compared against a keyframe threshold distance. If the threshold is exceeded, a new keyframe view $\symKfVertex$ is created from $\symAgentVertex$ and added to the keyframe set $\symKfVertexSet$, with an edge connection to the previous keyframe to ensure a connected path to the home location is always maintained.

\subsection{View Analysis and Coverage Sampling} \label{sec:approach:viewSampling}

$\symNbvVertexSet$ represents the set of NBV subgoal candidates expected to observe currently unknown voxels, such that map coverage will be increased if a subgoal is visited by the robot. To support the purposes of non-myopic planning, $\symNbvVertexSet$ should be sufficiently distributed to provide maximum coverage of the unknown map space. Additionally, maximum coverage should be achieved using a minimal size of $\symNbvVertexSet$ to reduce the eventual planning complexity that can increase exponentially with the number of views considered.

A sampling-based approach is used to incrementally build $\symNbvVertexSet$ to maintain maximum coverage as the map evolves. To efficiently and scalably achieve the aforementioned characteristics desired of $\symNbvVertexSet$, we introduce an approach using a frontier-based heuristic to evaluate information gain and also guide the sampling of additional views. 

\parName{\textbf{Information gain analysis:}}
A common approach in the literature to evaluate the expected information gain of a viewpose is by tracing the voxels along a dense set of raycasts within view's FoV, projected from its origin. This has a high computational cost that can become prohibitive when evaluating many views and as the map resolution increases. Additionally, it is difficult to efficiently determine the visible information overlap between different views, such that information gain is usually treated as an independent measure between views. This prevents an understanding of the unique or redundant coverage within a set of views, or how efficiently they cover the given map.

To mitigate these drawbacks, we directly use the frontier voxels within a view's FoV to constrain the evaluation of information gain. Given a voxel along a ray is only considered visible if no occupied voxels precede it, it can be inferred that the first unknown voxel traversed by a ray must be preceded by a free voxel to satisfy the visibility conditions. This transition from a free to an unknown voxel natural represents a frontier boundary, allowing a precondition to be defined that any raycast capable of containing information gain must at some point cross a frontier boundary. This allows the subset of raycasts that may contain some gain to be quickly identified based on the visible frontiers, which can greatly reduce the number of discrete raycast operations considered per view.

A \textit{visibility map} $\symViewVisMap : \symNodeSet \rightarrow \symFrontierSet$ is used to store the visible frontier features of each viewpose:

\begin{equation}\begin{gathered}
    \symViewVisMap( \symView ) = \{ \symFrontier \in \symFrontierSet  : \symVisFn( \symVoxel_{\symFrontier}, \symView ) \},
\end{gathered}\end{equation}

\noindent where $\symVoxel_{\symFrontier}$ is the voxel associated to $\symFrontier$, and $\symVisFn$ is an indicator function returning true if $\symVoxel_{\symFrontier}$ is visible from $\symView$. An \textit{inverse visibility map} $\symFrontierVisMap : \symFrontierSet \rightarrow \symNodeSet$ represents the preimage of $\symViewVisMap$ storing the viewposes from which each frontier is visible as 

\begin{equation}\begin{gathered}
    \symFrontierVisMap(\symFrontier) = \{ \symView \in \symViewSet  : \symVisFn( \symVoxel_{\symFrontier}, \symView ) \}.
\end{gathered}\end{equation}

The \textit{individual gain}, $\symIndividualGain$, of a view $\symView$ refers to the independent amount of unknown space visible from the view. This measure can be lower bounded by the number of visible frontiers $\symIndividualGain : \symViewSet \mapsto \lvert \symViewVisMap(\symView) \rvert$, since each frontier corresponds to an unknown voxel location. The \textit{joint gain}, $\symJointGain$, refers to the unique information collectively visible from a set of views. These can be respectively formulated as follows: 

\begin{equation}\begin{gathered}
    \symIndividualGain( \symView ) = \lvert \symViewVisMap(\symView) \rvert,
\end{gathered}\end{equation}

\begin{equation}\begin{gathered}
    \symJointGain( \symViewSet ) = \lvert \bigcup_{\symView \in \symViewSet}  \symViewVisMap(\symView)  \rvert.
\end{gathered}\end{equation}

The \textit{exclusive gain}, $\symExclusiveGain$, of a view $\symView$ refers to its unique contribution to the joint gain, or, in other words, the exclusively information visible by $\symView$ that is not visible by any other view in $\symViewSet$. $\symExclusiveGain$ can be determined according to the visible frontiers of $\symViewVisMap( \symView )$ that are only observed by $\symView$. This can be efficiently computed in linear time on the number of visible frontiers by:

\begin{equation}\begin{gathered}
    \symExclusiveGain( \symView ) = \lvert \{ \symFrontier \in \symViewVisMap(\symView) \ : \ \lvert \symFrontierVisMap(\symFrontier) \rvert = 1 \} \rvert.
\end{gathered}\end{equation}

\parName{\textbf{Coverage view sampling:}}
An iterative objective of DFR is to ensure maximum coverage of the current unknown space is maintained. $\symFrontierVisMap$ supports evaluation of the coverage completeness of the unknown map space by the current views $\symViewSet$. Let $\symFrontierSetCovered$ represent the set of covered frontiers, where a frontier is considered covered if it has at least one covering view able to observe it according to $\symFrontierVisMap$. The residual set is represented as $\symFrontierSetNoncovered = \symFrontierSet \setminus \symFrontierSetCovered$, and the global coverage completeness is evaluated by the fraction of covered frontiers, $\symFrontierSetCovered / \symFrontierSet$. The iterative coverage maximization objective can be formulated as:

\begin{equation} \label{eq:maxCoverageObj}
\begin{split} 
\max \frac{ \lvert \symFrontierSetCovered \rvert}{\lvert \symFrontierSet \rvert}
= \max \lvert \bigcup_{\symFrontier \in \symFrontierSet} \{ \symFrontier \mid \exists \symView \in \symViewSet, \symVisFn(\symVoxel_{\symFrontier}, \symView ) \} \rvert
.
\end{split}\end{equation}

\DefineMathCmd{\symFrontierQ}{\widehat{\symFrontierSet}}
\DefineMathCmd{\symfi}{\symFrontier_{i}}
\DefineMathCmd{\symvi}{\symView_{j}}
\DefineMathCmd{\symfVis}{\symFrontierSet_{\symvi}^{vis}}
\DefineMathCmd{\SYMSamplingVol}{\ensuremath{B_{\symFrontier_{i}}}}

A frontier-guided sampling strategy is presented to perform the maximization of (\ref{eq:maxCoverageObj}) by iteratively sampling viewposes to observe the non-covered frontiers. This effort is concentrated within $\symLocalDiffVolume$ which contains the most recent changes to the frontier distribution. Given the high complexity potentially involved in the sampling procedure, a performance tuning parameter $\symSamplingFactorLocal \in (0,1]$ is provided, representing a probability threshold used to select a random subset of the frontiers in $\symLocalDiffVolume$ to be considered for sampling in the current cycle.

A second parameter $\symSamplingFactorGlobal \in (0,1]$ is provided which serves a similar purpose as $\symSamplingFactorLocal$, but is applied to any non-covered frontiers that lie outside of $\symLocalDiffVolume$. This is to account for possible frontiers that were not successfully covered in a finite number of attempts during previous DFR cycles, which can result when large amounts of occupied or unknown space exist near a frontier. The difficulty in finding a feasible viewpose can greatly increase for these cases, and in some cases one may not exist with the available map knowledge. Given the increased difficulty, $\symSamplingFactorGlobal$ is given a lesser value than $\symSamplingFactorLocal$, allowing the search effort to persist between DFR cycles but with lower priority. In effect, this offers a degree of probabilistic completeness as the likelihood of finding a valid sample, if one exists, can continually increase over time while reducing the individual search effort per DFR cycle.

The sampling procedure is given in Alg. \ref{alg:ig_maximization}, which begins by calling $reconditionVisibility$ to update the visible information of existing views withing the changed volume. Between cycles, the frontier boundaries are often pushed back by only a small amount, but remain visible within the many of the same view as the previous cycle. This step ensures these differences are updated, so sampling is only needed when frontiers are pushed beyond visibility of all existing views.

Next, a frontier queue $\symFrontierQ$ is initialized containing the selected subsets from $\symFrontierSetNoncovered$. For each $\symfi \in \symFrontierQ$, a sampling subspace $\SYMSamplingVol$ is computed from which $\symfi$ can potentially be observed given the sensing parameters. For a maximum of $\symFrontierSampleThresh$ attempts, viewposes are randomly sampled using $getCoverageSample$ and checked by $isValidSample$ to determine if a valid sample has been found. A sample is considered valid only if it is collision-free and successfully observes the current frontier target, $\symfi$. 

Upon finding a valid sample, it is used to add a new node to the network, and all of its visible frontiers are computed to update the visibility map. If any of these frontiers are contained in $\symFrontierQ$, they are removed since they have been already covered by the current sample. This can greatly reduce the number of samples, since in practice a single view will often be able to observe many nearby frontiers.

\begin{algorithm}[tbp]
\caption{Frontier-guided view sampling for information gain maximization}
\label{alg:ig_maximization}
$ reconditionVisibility(\symGraph, \symLocalDiffVolume) $ \;
$ \symFrontierQ \gets frontierQueueInit (\symFrontierSetNoncovered, \symLocalDiffVolume, \symSamplingFactorLocal, \symSamplingFactorGlobal) $ \;
\While{ $ \symFrontierQ \neq \emptyset $ } {
    $ \symfi \gets extractNext ( \symFrontierQ ) $\;
    $ \SYMSamplingVol \gets getSamplingVol(\symfi,\symSensorRange, \symFov) $\;
    $ success \gets false, n \gets 0 $ \;
    \While{ $ n < \symSamplingMaxAttempts \And \neg success $ } {
        $ \symPoseSample \gets getCoverageSample( \SYMSamplingVol ) $\;
        \If{ $ isValidSample (\symPoseSample) $ }{
            $ \symvi = addNode ( \symPoseSample, \symGraph) $\;
            $ \symfVis \gets computeVisible (\symvi, \symfi) $\;
            $ updateVisibility ( \symvi , \symfVis ) $\;
            $ \symFrontierQ = \symFrontierQ \setminus \symfVis$\;
            $ success \gets true $ \;
        }
        $n = n + 1$ \;
	}
}
\end{algorithm}

\subsection{Pruning and Refinement} \label{sec:approach:viewRefinement}

The growth rate of the network is reduced by pruning unnecessary views that no longer provide any individual gain contribution, and redundant views with little or no exclusive information gain. These conditions naturally occur as the robot progresses its exploration of the map and observes the previously unknown space within each view. They also occur as a result when new view samples are added to the network which overlap with the pre-existing views, decreasing their exclusive gain. The goal is to identify the views that can be removed from the network without loss of the overall joint gain.

The joint gain and exclusive gain measures are used to formulate the \textit{pruning objective} as a submodularity maximization problem. Given an initial set of views $\symViewSet$, pruning can be described as finding the minimum subset of views $\symViewSet^{*}$ that achieves the same total joint gain as $\symViewSet$, as follows:

\begin{equation} \label{eq:pruningObjective}
\begin{gathered}
\underset{ \symViewSet^{*} \subseteq \symViewSet } {\text{argmin}}( \symViewSet^{*} ), \\
\text{s.t.} \\
\symJointGain(\symViewSet) - \symJointGain( \symViewSet^{*} )  \approx  0.
\end{gathered}
\end{equation}

To solve (\ref{eq:pruningObjective}), a set of pruning candidates is found by searching for views that have negligible individual or exclusive information gain. Given the local difference neighborhood $\symLocalDiffVolume$, the search is restricted to the views located within visible range $\symSensorRange$ of $\symLocalDiffVolume$, corresponding to the views with visibility information that was potentially effected by the map changes. The candidates within this region are further evaluated for their edge connectivity. Any candidate found to have a cut-edge is preserved to maintain the graph connectivity, while the remainder are deleted.

Once the pruning stage is complete, the coverage views of $\symNbvVertexSet$ represent the supremal set that maximizes map coverage using a minimal number of views. Not only does this help to reduce the total size, but the minimization of redundant coverage helps to simplify the planning problem. Since each NBV has some positive amount of exclusive gain after pruning, they represent an exact set of targets that a planner must determine how to optimally visit, without the need to evaluate their redundancy during its search.

\subsection{Reachability Update} \label{sec:approach:traversalEdges}

\DefineMathCmd{\symEdgePairCandidates}{\symEdgeSet_{local}}
\DefineMathCmd{\tsymEdgePairCandidate}{\symEdge_{i}}

The reachability knowledge represented by $\symEdgeSet$ is updated each iteration to account for new map knowledge and any state changes in $\symApnVertexSet$. Additional nodes are also sampled during this stage to increase the overall node density and uniformity in  $\symTraversalVertexSet$. This accounts for the non-uniformity of coverage view sampling, which is biased towards the frontier boundaries. Since the purpose of $\symTraversalVertexSet$ is primarily to increase the network connectivity, only the position of these samples is needed, while the visible information and pose orientation attributes can be ignored.

\begin{algorithm}[tbp]
\caption{Reachability expansion algorithm.}
\label{alg:reach:edgeMaximization}
$ \symReachabilityUpdateVol \gets getSearchVolume(\symLocalDiffVolume, \symMinTraversalSepRadius) $ \;
\tcp*[l]{Stage 1: increase node density }
$ i, n \gets 0$\;
\While{ $ i < \symMaxTraversalSampleAttempts \And n < \symMaxTraversalSamples  $ } {
    $ \symPos_{i} \gets generateReachabilitySample(\symReachabilityUpdateVol) $ \;
    $ \symPos_{near} \gets findNearest(\symPos_{i}, \symViewSet) $ \;
    \If{ $ distance(\symPos_{i}, \symPos_{near}) > \symMinTraversalSepRadius $}{
        $ \fnAddNode( \symAPN, \symPos_{i} ) $\;
        $ n = n + 1$\;
	}
    $ i = i + 1$\;
}
\tcp*[l]{Stage 2: increase edge density}
$ \symEdgePairCandidates \gets getUncertainEdgePairs ( \symReachabilityUpdateVol, \symEdgeUpdateProbThresh ) $ \;
\For{ $ \tsymEdgePairCandidate \in \symEdgePairCandidates $ }{
    $ \symEdgeBoundingVol, \symEdgeCollisionState, \symEdgeObsPoint \gets computeEdgeState(\tsymEdgePairCandidate) $ \;
    \uIf{ \symEdgeCollisionState = free } {
        $ \fnAddEdge(\symAPN, \tsymEdgePairCandidate) $ \;
    }
    \uElseIf{\symEdgeCollisionState = unk} {
        $ \fnAddUnknownEdge(\symAPN, \tsymEdgePairCandidate, \symEdgeBoundingVol, \symEdgeCollisionState, \symEdgeObsPoint) $ \;
    } 
    \ElseIf{\symEdgeCollisionState = occ } {
        $ \fnAddCollisionEdge(\symAPN, \tsymEdgePairCandidate) $ \;
    }
}
\end{algorithm}

The pseudocode for the reachability update procedure is shown in Alg. \ref{alg:reach:edgeMaximization}, which contains two primary stages. The first stage samples traversal nodes to increase the distribution density within the graph, and the second stage increases the total edge density. 

In the first stage, collision-free positions are uniformly sampled from $\symReachabilityUpdateVol$, for a maximum of $\symMaxTraversalSampleAttempts$ attempts, or until a threshold of $\symMaxTraversalSamples$ samples are accepted. Each sample is evaluated according to the distance of its nearest neighbor in $\symViewSet$, and compared against a threshold distance, $\symMinTraversalSepRadius$. $\symMinTraversalSepRadius$ serves as a density constraint to prevent too many samples from being added in close proximity, which would unnecessarily increase the size complexity of the graph while adding little or no additional reachability knowledge. A sample is accepted if its nearest neighbor distance is greater than $\symMinTraversalSepRadius$, and a new node is added to the graph using the sampled position.

The second stage begins by extracting the local set of candidate edge pairs $\symEdgePairCandidates$ using the function $getUnknownEdgePairs$. This procedure searches $\symReachabilityUpdateVol$ to find the set of node pairs $( \symApnVertex_{u}, \symApnVertex_{w} )$ such that the collision state of the corresponding edge $\symEdge_{u,w}$ is either null or unknown. Here, a null edge indicates the edge does not exist (i.e. has not been evaluated in any DFR cycle), while unknown refers to an edge found with an uncertain collision state from a previous DFR cycle. A parameter $\symEdgeUpdateProbThresh \in [0,1)$ is used to specify a random probability threshold of whether to evaluate a candidate node pair $( \symApnVertex_{u}, \symApnVertex_{w} )$. This helps to limit the number of edge evaluation operations that occur per cycle, similar the parameter $\symSamplingFactorLocal$ used for coverage view sampling.

Each edge is evaluated by $computeEdgeState$ to determine its collision state data, which leverages previously cached results if available. Since edges may be evaluated between any nodes over any distance within $\symReachabilityUpdateVol$, the cached collision data can significantly reduce the update complexity. If an occupied collision is found, the edge is added to the cache of collision edges to prevent future evaluation. For unknown voxel collisions, the edge is added to the cache of uncertain edges along with the intermediate collision data results to accelerate future re-evaluation. Otherwise, the edge is added to the graph by $addEdge$ which computes and stores its associated cost information according to (\ref{eq:edgeCost}) for efficient lookup by other procedures and planning.

\subsection{Topological Clustering} \label{sec:approach:clustering}

The graph nodes are decomposed into a set of subgraph regions represented by the hyperedges $\symHyperEdgeset$, as illustrated in Fig. \ref{f:approach:APNClusterGraph}. $\symHyperEdgeset$ serves as a topological hierarchy over $\symGraph$ to reduce its size complexity. This representation can be utilized to increase the efficiency for search, traversal, and other operations. A tradeoff occurs where greater reductions in size complexity also result in reduced level of detail (LoD), i.e. resolution.

To compute the hyperedges, we use a density-based clustering approach based on \cite{ester1996density, schubert2017dbscan}, extended to leverage both the geometric and reachability knowledge already present in the APN. The algorithm uses two parameters, $\clusteringEps$ and $\clusteringMinPts$, where $\clusteringEps$ defines neighborhood distance threshold, and $\clusteringMinPts$ defines a density threshold for the neighborhood. 

Let a node $\symApnVertex_{p}$ be defined as a \textit{core node} if it has at least $\clusteringMinPts$ edge-connected neighbors within distance $\clusteringEps$. A node $\symApnVertex_{q}$ is then defined as a \textit{reachable node} from $\symApnVertex_{p}$ only if there exists an edge connection between $\symApnVertex_{p}$ and $\symApnVertex_{q}$, and $\symApnVertex_{q}$ is within distance $\clusteringEps$ from $\symApnVertex_{p}$. Given a core node $\symApnVertex_{p}$, a cluster is formed by all nodes reachable from $\symApnVertex_{p}$. Any remaining nodes that are neither core nodes nor reachable from a core node are assigned as singleton clusters.

This approach allows clusters to form more naturally by additionally considering the edge connectivity between points. They are also not required to be geometrically convex as with other clustering approaches. This enables fewer clusters to be formed, since they can be better fit to the nodes over arbitrarily shaped space. Explicit constraints on the maximum number of clusters or their size are also not necessary, such that clusters can conform to the map with variable size and density, which can effectively handle environments where different regions may have different geometric characteristics and complexities.

\section{Hierarchical Evolutionary View Planning} \label{sec:approach:viewPlanning}

\DefineMathCmd{\clusterIndexSet}{\mathbb{J}^{\symHypergraph}}
\DefineMathCmd{\viewIndexSet}{\mathbb{J}^{\symViewSet}}
\DefineMathCmd{\symSymGroup}{\boldsymbol{S}_{n}}
\DefineMathCmd{\clusterIndexSetSymGroup}{\boldsymbol{S}_{n}}

\begin{figure}[tbp]
\centering
\includegraphics[width=0.97\columnwidth, trim={0 0 0 0 },clip]{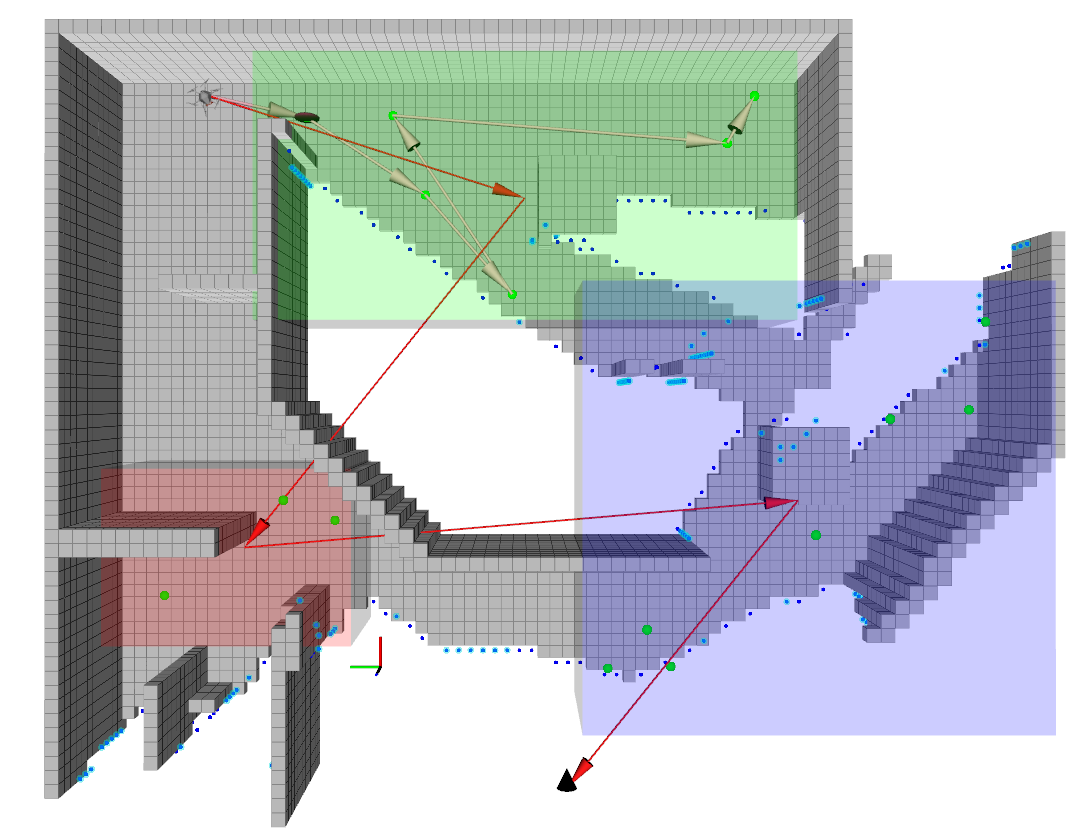}
\caption{Visual depiction of the hierarchical planning strategy. The first stage computes the global path (dark red arrows) over the node clusters, with the start fixed to the robot location and the end fixed to the home location. The second stage optimizes the NBV sequence (depicted using tan arrows) within the first cluster of the global sequence (green bounding box).
}
\label{f:approach:planningDepiction}
\end{figure}

The iteratively updated APN provides a generalized representation of the exploration state space, which can be utilized by any graph-based planning strategy for global and local planning. In this section, we present an anytime planning approach referred to as the APN Planner (APN-P), which leverages the hierarchical decomposition of the APN to plan a global sequence over the topological subgraph regions. A second planning stage then optimizes the low-level view path for the first subgraph of the topological sequence. Each stage is formulated as a Fixed-Ended Open Traveling Salesman Problem (FEOTSP), solved using an evolutionary optimization approach to determine the optimal sequence orders. A visualization of this procedure is displayed in Fig. \ref{f:approach:planningDepiction}. 

Let $\clusterIndexSet \subset \mathbb{N}$ be an index set that enumerates the clusters $\symViewClusterSet$. A cost matrix $\symClusterCostMatrix$ is computed by finding the shortest path between the centroid views of each pair of clusters. Given the pairwise cost $s(u, w) \in \symClusterCostMatrix$ between cluster indices $u, w \in \clusterIndexSet$, the cluster planning objective is to find the minimum cost permutation $\symViewClusterPath \in \symSymGroup(\clusterIndexSet)$ of the indices $\clusterIndexSet$, where $\symSymGroup(\clusterIndexSet)$ is the symmetric group of $\clusterIndexSet$. 

Given first cluster $\symViewCluster_0$ of $\symViewClusterPath$, and its induced subgraph $\symGraph [ \symViewCluster_{0} ]$, the view planning procedure is similarly formulated. Given an index set $\viewIndexSet \subset \mathbb{N}$ enumerating the NBVs $\{\symView\} \in \symGraph [ \symViewCluster_{0} ]$, a pairwise cost matrix $\symViewpointCostMatrix$ between index pairs $u, w \in \viewIndexSet$ can be obtained directly from the existing edge costs. The view path planning objective is then to find the minimum cost permutation $\symViewPath = ( \symAgentNode, \symView_{\viewIndexSet_{0}}, \cdots, \symView_{\viewIndexSet_{n}} )$, which begins at the current robot configuration $\symAgentNode$ and visits each NBV node of the target cluster.

The optimized sequences are preserved in a data cache allowing them to be used to re-initialize subsequent planning cycles. Given a planning cycle $i$ and target cluster $ \symGraph [ \symViewCluster_{0} ] $, the current solution $\symOrderedPath_{i}$ is initialized from $\symOrderedPath_{i-1}$ by first filtering out any invalid views $ \symOrderedPath_{i-1} \setminus \symGraph [ \symViewCluster_{0} ]$ that do not belong to the current cluster. The relative order over the common subset $ \symOrderedPath_{i-1} \cup \symGraph [ \symViewCluster_{0} ]$ is preserved, and any additional views $\symGraph [ \symViewCluster_{0} ] \setminus \symOrderedPath_{i-1} $ are inserted using local search to estimate their optimal sequence positions.

Sequence optimization is performed using a mimetic evolutionary algorithm \cite{deb2002fast}. A population of $P_{n}$ candidates, or individuals, are initialized by randomized permutations of $\hat{\symOrderedPath}$. For a maximum of $N_{g}$ generations, the population is optimized using a pairwise swap mutation and partially mapped crossover (PMX) \cite{kora2017crossover}. The procedure terminates once $N_{g}$ generations is exceeded, or an improved solution cannot be found after $N_{stall}$ generations. 

Once the exploration plan optimization is complete, the first view of the local sequence represents the navigation goal, $\symView_{g}$. If this goal is different from the previous goal, the cost of their respective sequences is compared to determine whether to accept or reject the new goal, penalizing significant changes in the direction of motion. Once the appropriate goal is selected, its trajectory is computed with $RRT^{*}$ \cite{sucan2012open}, using the APN to find the shortest path to initialize the trajectory planner. Exploration terminates once no frontiers remain, or no reachable views can be found for the remaining frontiers.

Given that solutions tend to become more optimal with more generations, the computational efficiency of DFR directly impacts the planning quality. This can have a compounding effect also, since the planning convergence rate can be increased with increased optimality of prior solutions used to initialize subsequent instances.

\section{Evaluation} \label{results}

\assign{\valSafetyRadius}{0.75}

\begin{figure*}[!htbp]
\centering
\captionsetup{justification=centering}
	\subcaptionbox{Apartment%
		\label{f:worldScenarios:flat:worldModel}}%
		{\includegraphics[width=0.3\textwidth]{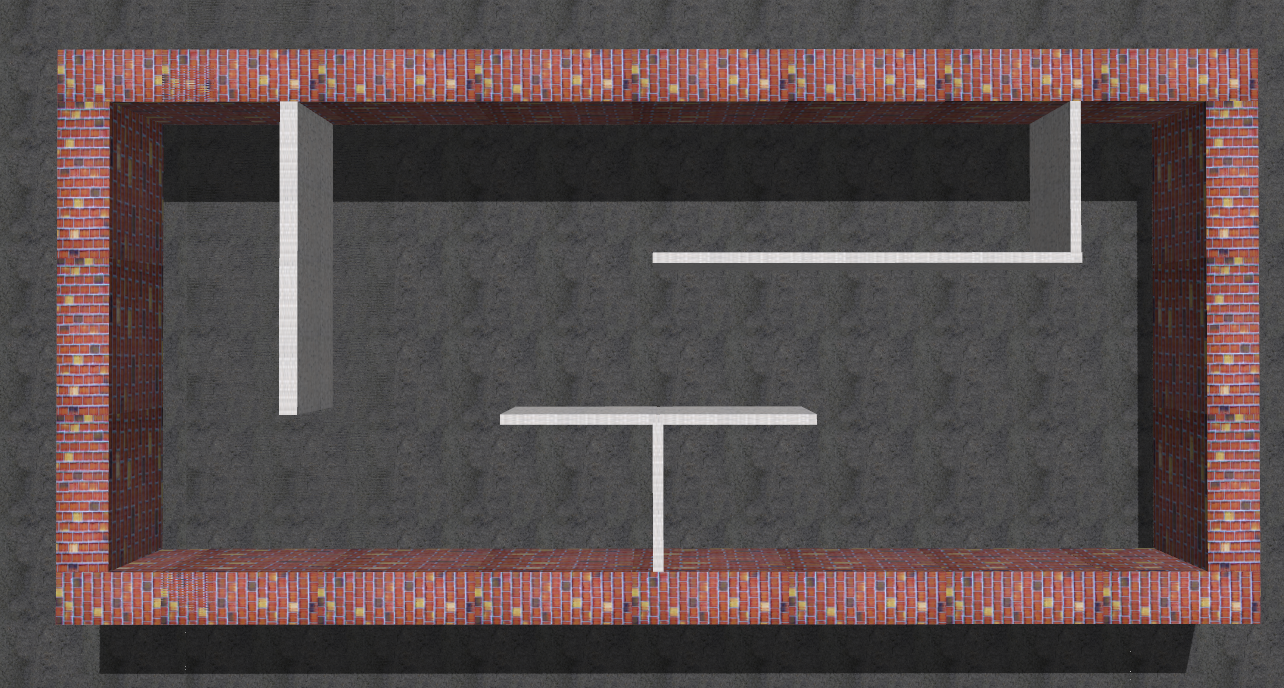}}%
	\qquad\hfill
	\subcaptionbox{Maze %
		\label{f:worldScenarios:maze:worldModel}}%
		{\includegraphics[width=0.3\textwidth]{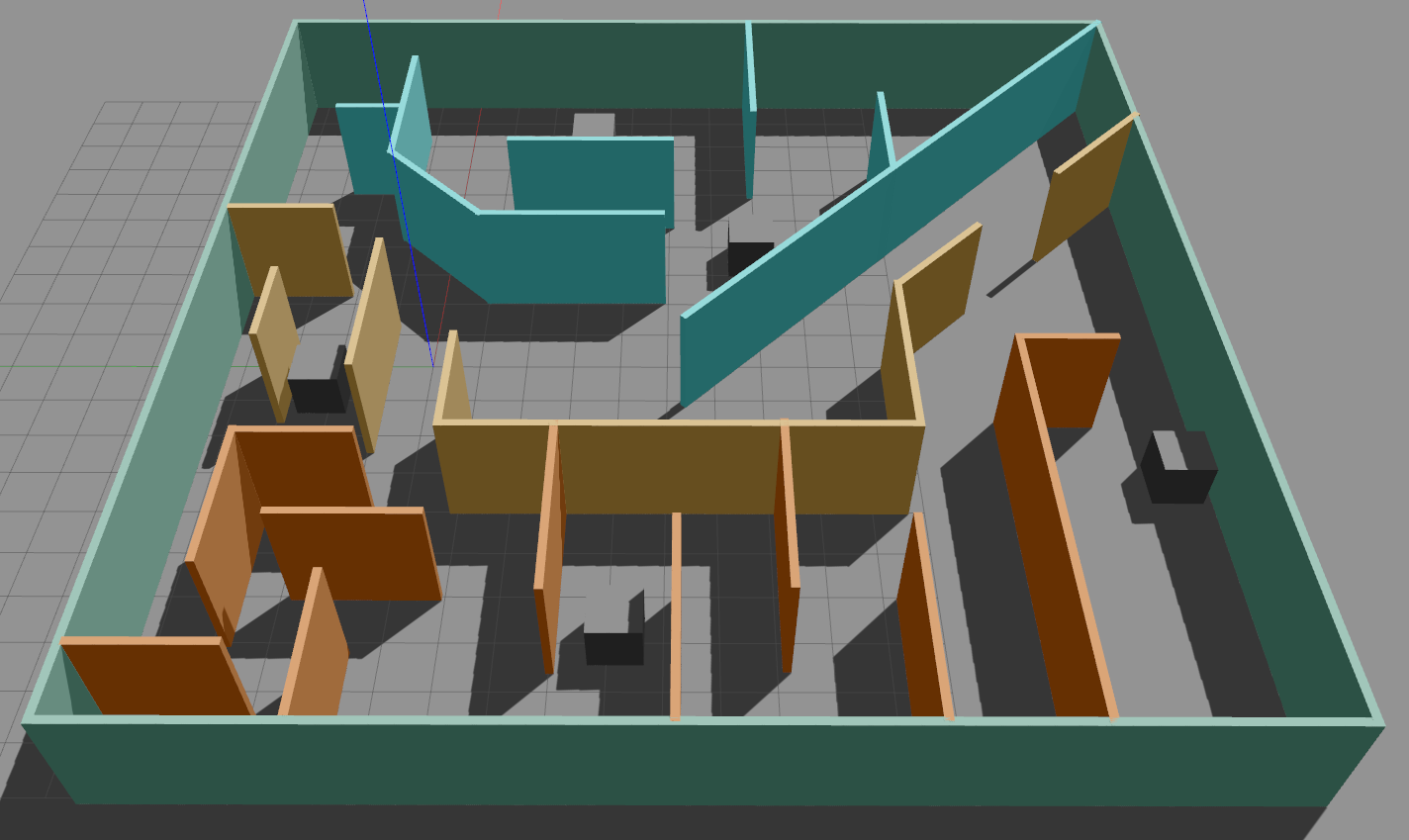}}%
	\qquad\hfill
	\subcaptionbox{Industrial Plant %
		\label{f:worldScenarios:plant:worldModel}}%
		{\includegraphics[width=0.3\textwidth]{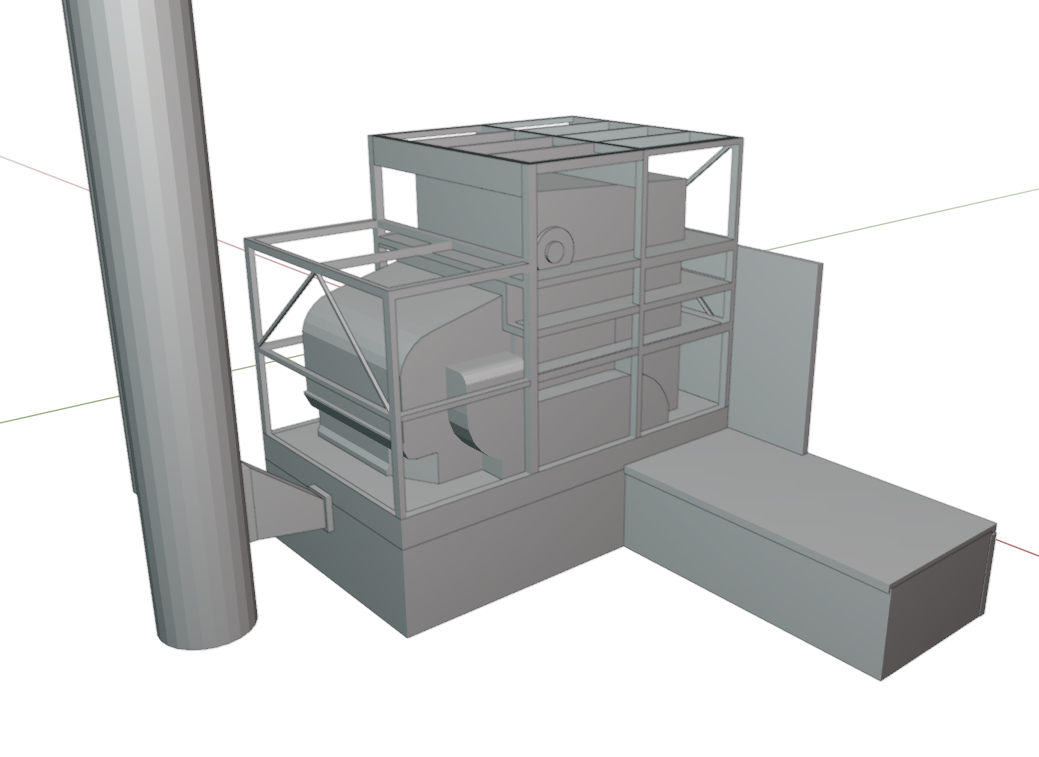}}%
	\\ \hfill
	\subcaptionbox{Warehouse %
		\label{f:worldScenarios:warehouse:worldModel}}%
		{\includegraphics[width=0.44\textwidth]{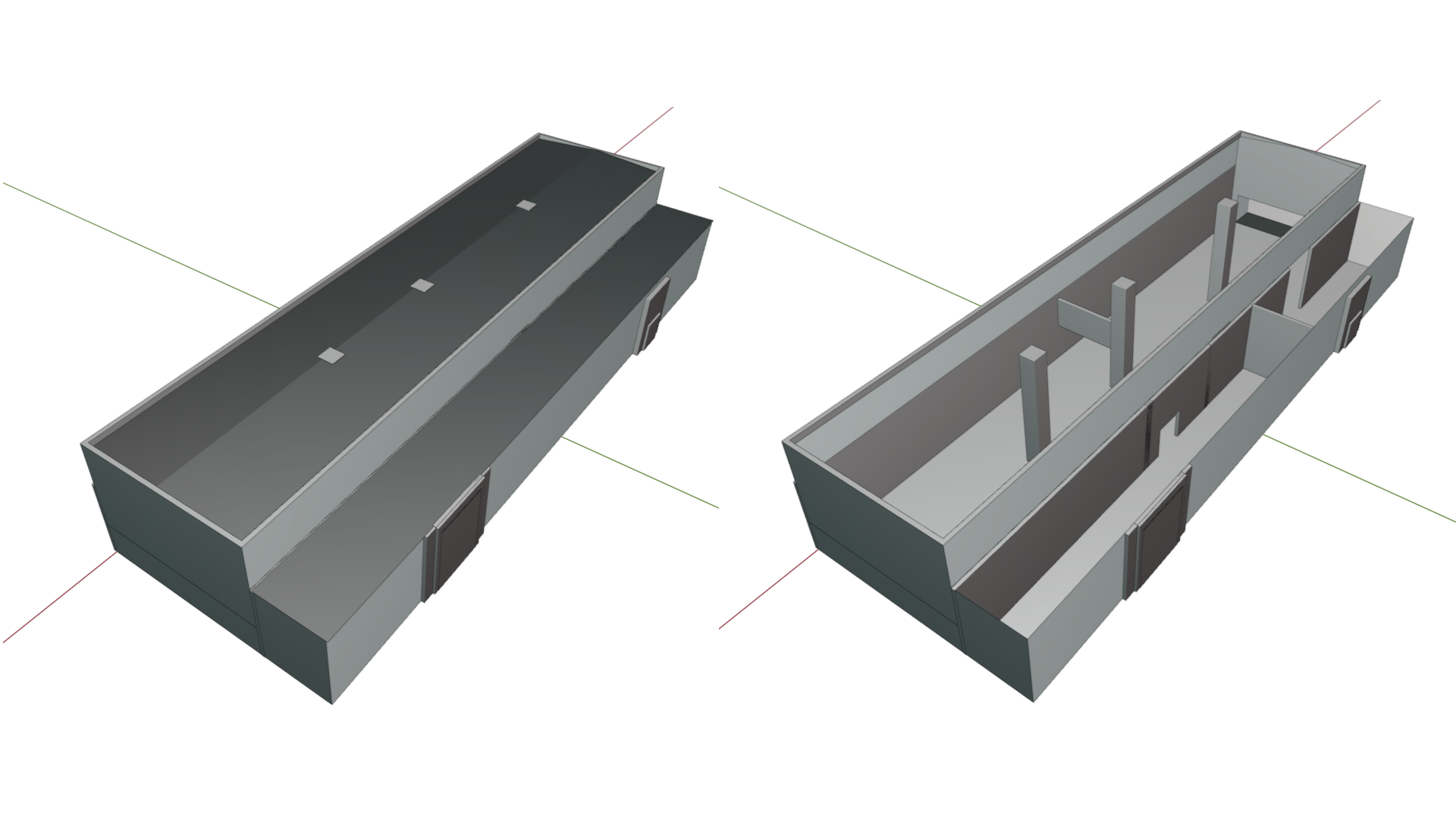}}%
	\qquad\hfill
	\subcaptionbox{Relative environment scale comparison %
		\label{fig:worldDimensions}}%
		{ \includegraphics[width=0.44\textwidth, trim={0 0cm 0 3cm}, clip]{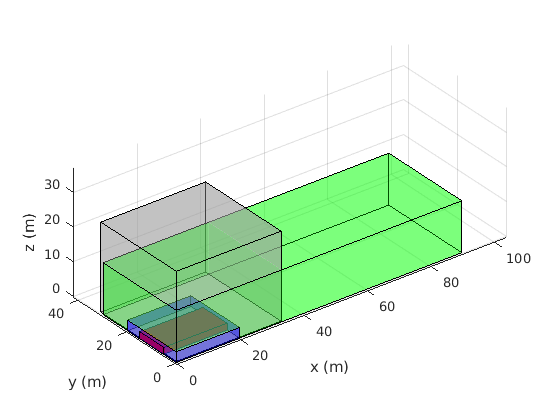} }%
	\qquad\hfill
\caption{Visualization of each evaluated world scenario. The relative scale of each scenario is depicted in \ref{fig:worldDimensions} according to their bounding box dimensions, where red represents the Apartment (slightly offset from the origin for visual clarity), blue represents the Maze, grey represents the Industrial Plant, and green represents the Warehouse.}
\label{f:worldScenarios}
\end{figure*}

The APN and APN-P were evaluated through ROS-based simulations using Gazebo \cite{koenig2004design} and the RotorS MAV simulation framework \cite{furrer2016rotors}. The AscTec Firefly MAV model provided by RotorS was used to simulate the robot dynamics and control systems, and was equipped with a stereo depth sensor for visual perception. The simulations and all algorithms were executed using a single laptop computer with Intel Core i7 2.6 GHz processor and 16 GB RAM. The test results were used to analyze the computational performance and planning efficiency of the proposed approach.

Exploration was tested using several different 3D structure models with various scales as displayed in Fig. \ref{f:worldScenarios}, with a visual comparison of their relative scales shown in Fig. \ref{fig:worldDimensions}. In addition to varying sizes, each environment provides different characteristics for evaluation, such as obstacle density, narrow spaces opposed to open space, dead-ends, and overall geometric complexity.

A video presentation demonstrating the operation and performance of the APN-P is included with this work. The simulation environment is used to visualize the concepts of operation as they are executed. The Apartment scenario is used in the video presentation to demonstrate the real-time operation of the full exploration procedure while visualizing the APN's dynamically changing structure.

To account for the stochastic nature of the approach, each scenario was run 5 times and statistical analysis was computed over a variety of performance metrics, summarized in Table \ref{t:eval:metrics}. The average total exploration runtime required to complete the exploration task is denoted as $\symRunTime$, and $\symCycleTime$ refers to the average computational time required per update and planning cycle. A maximum exploration time limit of $T_{max} =  14\text{min}$ ($840\Sec$) was imposed, which is the maximum rated flight time for the AscTec Firefly. If this threshold is exceeded, exploration immediately terminates and failure is reported.

\renewcommand{\arraystretch}{1.3}
\begin{table}[bt]
    \footnotesize
    \centering
    \caption{Summary of performance analysis metrics.}
\begin{tabular}{|m{0.11\columnwidth}| m{0.78\columnwidth} |}
\hline
    Symbol & Description
\\ \hline
    $ \symRunTime$ & average total exploration runtime ($\unitsSeconds$) 
\\ \hline
    $ \symCycleTime$ & average computational time per cycle  ($\unitsMillisec$)
\\ \hline
    $\symCoverageRatio$ & final map surface coverage ratio ($\%$) as $\symVoxelSetOcc /\symVoxelSetOccFullCoverage$
\\ \hline
    $\symVolumetricRate$ & average voxel discovery rate 
 ($\CubicMetersPerSec$)
\\ \hline
    $\symSurfaceExploreRate$ & average surface voxel discovery rate  ($\CubicMetersPerSec$)
\\ \hline
    $\symNodeDensityFactor$ & avg. number of nodes per unit of map volume  ($1 / 100 \unitsMetersCubed$)
\\ \hline
    $\symEdgeDensityFactor$ & avg. ratio of known edges $\lvert \symEdgeSet \rvert $ to possible edges $ \binom{\lvert \symApnVertexSet \rvert}{2}$
\\ \hline
\end{tabular}
\label{t:eval:metrics}
\end{table}
\renewcommand{\arraystretch}{1}

\renewcommand{\arraystretch}{1.3}
\begin{table}[tbp]
    \footnotesize
    \centering
    \caption{Summary of common configuration parameters.}
    \label{t:apartment:params}
\begin{tabular}{| c | cccc |}
\hline
    &  \multicolumn{4}{c|}{Scenario} \\
    \hline
    Param & Apt. & Maze & Ind. Plant &Warehouse
\\ \hline
    $ \symMapRes $ & $\{ 0.1,0.2,0.4  \}$ & $\{ 0.1,0.2 \}$ & $\{ 0.2 \}$ & $\{ 0.4 \}$
\\ \hline
    $\symSafetyRadius$ & \multicolumn{4}{c|}{$\valSafetyRadius \unitsMeters$}
\\ \hline
    $\symMaxVel $ & $1.0\unitsVelocity$ & $2.0\unitsVelocity$ & $2.5\unitsVelocity$ & $3.0\unitsVelocity$
\\ \hline
    $ \symMaxYaw $ & \multicolumn{4}{c|}{$0.75\unitsAngVelocity$} 
\\ \hline
    $ \symSensorRange $ & $5\unitsMeters$ & $6\unitsMeters$ & $7\unitsMeters$ &$ 9\unitsMeters$
\\ \hline
    $ \symVFov, \symHFov $ &  [60\textdegree, 90\textdegree] & [60\textdegree, 90\textdegree] & [75\degree, 115\degree] & [75\degree, 115\degree]
\\ \hline
\end{tabular}\end{table}
\renewcommand{\arraystretch}{1}

The total map coverage is given as the ratio $\symCoverageRatio$ of the number of surface voxels $\symVoxelSetOcc$ discovered during exploration with respect to a ground truth set $\symVoxelSetOccFullCoverage$ of all visible surface voxels. $\symVoxelSetOccFullCoverage$ was determined by manually guiding the robot through each world scenario, carefully ensuring every observable surface was covered by the sensor. The total volumetric exploration rate is given as $\symVolumetricRate$, which is the average volume of new information gain per second in $\CubicMetersPerSec$. Since the objective is to achieve complete surface coverage, a more useful metric is $\symSurfaceExploreRate$ which refers to the rate of occupied information gain in $\CubicMetersPerSec$.

The APN is evaluated according to its average node density $\symNodeDensityFactor$ and edge density $\symEdgeDensityFactor$. Here, node density refers to the number of nodes within a standard unit of volume, normalized as the number of nodes per $100 \unitsMetersCubed$ of the mapped free space. Edge density refers to the ratio between the known edges $\lvert \symEdgeSet \rvert $ and the total edge capacity of a complete edge set over the nodes, $ \binom{\lvert \symApnVertexSet \rvert}{2}$. $\symNodeDensityFactor$ and $\symEdgeDensityFactor$ are given as the average over all cycles of the test scenario.

The following baseline approaches were used for comparative analysis with the APN-P:

\begin{itemize}
\item RH-NBVP \cite{bircher2018receding}: A receding horizon method that finds informative view paths using RRT-based expansion within a local region of the robot.

\item AEP \cite{selin2019efficient}: An approach that extends the strategy of RH-NBVP, using RH-NBVP for local planning and frontier-based planning for global search when local planning fails to find informative views.

\item FFI \cite{dai2020fast}: A hybrid frontier-based and sampling-based approach that uses an efficient frontier clustering strategy to guide the sampling of views.

\item Rapid \cite{cieslewski2017rapid}: An extension of frontier-based planning designed to maintain the fastest allowable velocity by guiding towards frontiers within the sensors current field of view, and using classical frontier planning when no visible frontiers are available.
\end{itemize}

A summary of common parameters for the different scenarios is shown in Table \ref{t:apartment:params}, which were selected as consistently as possible to the baseline approaches. The map resolution $\symMapRes$ was varied between the values $\{ 0.1, 0.2, 0.4 \} \unitsMeters$ to analyze its effects on performance scalability. The maximum linear velocity $\symMaxVel$ and yaw rate $\symMaxYaw$ were assigned based on the common values used in the comparative approaches, along with the sensing parameters $\symSensorRange$ and $( \symVFov, \symHFov )$. 

Coverage view sampling parameters related to Alg. \ref{alg:ig_maximization} were set as $\symSamplingFactorLocal = 0.8$, $\symSamplingFactorGlobal = 0.1$, and $\symSamplingMaxAttempts = 30$ for each scenario. The reachability update parameters for Alg. \ref{alg:reach:edgeMaximization} for each scenario were commonly set to $\symMaxTraversalSamples = 3$,  $\symEdgeUpdateProbThresh = 0.7$, and $\symMinTraversalSepRadius = 2.0 \unitsMeters$.

\newcommand{\apnpRes}[3]{
\def\testArgA{#1}\def\testArgB{#2}\def\testArgC{#3}%
\def\testTotalTime{totalTime}%
\def\testCycleTime{cycleTime}%
\def\testTotalDist{totalDist}%
\def\testAvgVel{AvgVel}%
\def\testAvgViewCount{AvgViewCount}%
\def\testAvgEdgeCount{AvgEdgeCount}%
\def\testEdgeDensity{EdgeDensity}%
\def\testCoverageRatio{coverageRatio}%
\def\testExploreRate{ExploreRate}%
\def\testSurfExploreRate{SurfExploreRate}%
\def\testFreeExploreRate{FreeExploreRate}%
\def\testApt{apt}%
\def\testMaze{maze}%
\def\testPlant{plant}%
\def\testWarehouse{warehouse}%
\def\testResIV{0.4}%
\def\testResII{0.2}%
\def\testResI{0.1}%
\ifx\testArgA\testApt
    \ifx\testArgC\testResIV
        \ifx\testArgB\testTotalTime
            52.9
        \fi
        \ifx\testArgB\testCycleTime
            \ensuremath{14.0 \pm 8.0}
        \fi
    \fi
    \ifx\testArgC\testResII
        \ifx\testArgB\testTotalTime
            57.9
        \fi
        \ifx\testArgB\testCycleTime
            \ensuremath{18.9 \pm 9.1}
        \fi
    \fi
    \ifx\testArgC\testResI
        \ifx\testArgB\testTotalTime
            69.4
        \fi
        \ifx\testArgB\testCycleTime
            \ensuremath{28.9 \pm 18.5}
        \fi
    \fi
\fi
\ifx\testArgA\testMaze
    \ifx\testArgC\testResII
        \ifx\testArgB\testTotalTime
            \ensuremath{ 145.1 }
        \fi
        \ifx\testArgB\testCycleTime
            \ensuremath{ 26.1 \pm 20.8 }
        \fi
    \fi
    \ifx\testArgC\testResI
        \ifx\testArgB\testTotalTime
            \ensuremath{ 212.6 }
        \fi
        \ifx\testArgB\testCycleTime
            \ensuremath{ 48.0 \pm 28.8 }
        \fi
    \fi
\fi
\ifx\testArgA\testPlant
    \ifx\testArgB\testTotalTime
        \ensuremath{ 353.1 }
    \fi
    \ifx\testArgB\testCycleTime
        \ensuremath{ 186.8  }
    \fi
    \ifx\testArgB\testCoverageRatio
        \ensuremath{ 98.7\% }
    \fi
    \ifx\testArgB\testTotalDist
        \ensuremath{ 406.3 }
    \fi
    \ifx\testArgB\testAvgVel
        \ensuremath{ 1.9 }
    \fi
    \ifx\testArgB\testAvgViewCount
        \ensuremath{198}
    \fi
    \ifx\testArgB\testAvgEdgeCount
        \ensuremath{3962}
    \fi
    \ifx\testArgB\testEdgeDensity
        \ensuremath{0.23}
    \fi
\fi
\ifx\testArgA\testWarehouse
    \ifx\testArgB\testTotalTime
        \ensuremath{ 268.1 }
    \fi
    \ifx\testArgB\testCycleTime
        \ensuremath{ 121.3 }
    \fi
    \ifx\testArgB\testCoverageRatio
        \ensuremath{99.98\%}
    \fi
    \ifx\testArgB\testExploreRate
        \ensuremath{9.2}
    \fi
    \ifx\testArgB\testSurfExploreRate
        \ensuremath{1.4}
    \fi
    \ifx\testArgB\testFreeExploreRate
        \ensuremath{7.8}
    \fi
    \ifx\testArgB\testTotalDist
        \ensuremath{645.1}
    \fi
    \ifx\testArgB\testAvgVel
        \ensuremath{2.4}
    \fi
    \ifx\testArgB\testAvgViewCount
        \ensuremath{346}
    \fi
    \ifx\testArgB\testAvgEdgeCount
        \ensuremath{21494}
    \fi
    \ifx\testArgB\testEdgeDensity
        \ensuremath{0.4}
    \fi
\fi
}


\assign{\mazeTotalTimeFFI}{-}
\assign{\mazeCycleTimeFFI}{-}

\assign{\mazeTotalTimeAEP}{-}
\assign{\mazeCycleTimeAEP}{-}

\assign{\mazeTotalTimeNBVP}{-}
\assign{\mazeCycleTimeNBVP}{-}

\assign{\mazeTotalTimeRapid}{-}
\assign{\mazeCycleTimeRapid}{-}

\assign{\plantTotalTimeFFI}{> 1000}
\assign{\plantCycleTimeFFI}{152 \pm 20}

\assign{\plantTotalTimeAEP}{941}
\assign{\plantCycleTimeAEP}{-}

\assign{\plantTotalTimeNBVP}{2104}
\assign{\plantCycleTimeNBVP}{-}

\assign{\plantTotalTimeRapid}{582}
\assign{\plantCycleTimeRapid}{-}


\subsection{Apartment Scenario}

\begin{figure}[tbp]
\centering
\captionsetup{justification=centering}
\subfloat[Reconstructed map and exploration path. \label{f:results:apartment:map}]{ 
	\includegraphics[width=0.85\columnwidth]{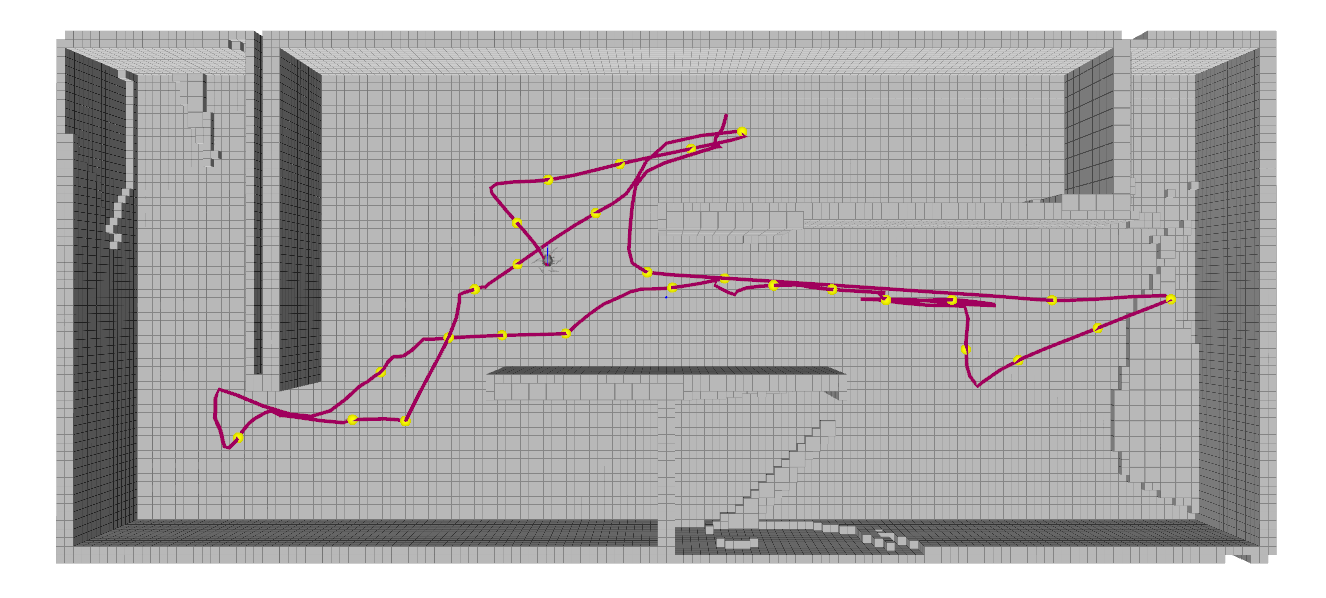}
    	\centering
}
\hfill
\subfloat[APN reachability roadmap. \label{f:results:apartment:apn}]{ 
	\includegraphics[width=0.85\columnwidth]{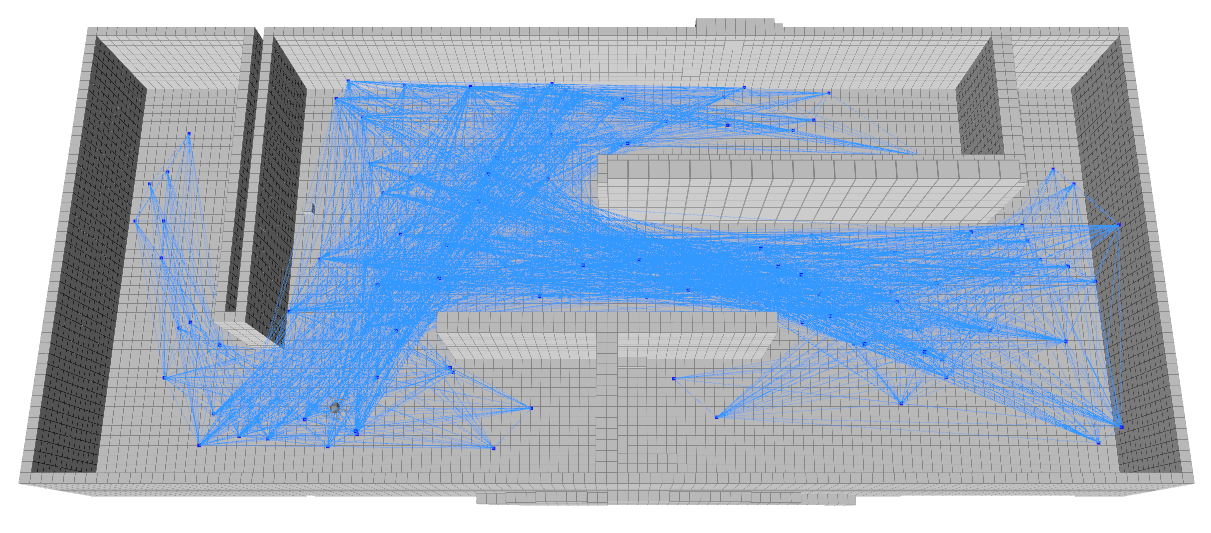}
	\centering
}
\caption{Exploration results for the Maze Scenario. (a): The explored path is plotted in red, with intermediate keyframe configurations represented by yellow points. (b): The APN nodes and edges overlayed in blue.}
\label{f:results:apartment}
\end{figure}

The apartment scenario in Fig. \ref{f:worldScenarios:flat:worldModel} is a relatively small scale interior space with the dimensions $20\times10\times 3(\unitsMetersCubed)$, used as a baseline for comparing the larger and more complex scenarios. An example map reconstruction by APN-P is shown in Fig. \ref{f:results:apartment:map} with the traced exploration path, and the APN roadmap is shown in Fig. \ref{f:results:apartment:apn}. The average distance traveled was $76.5 \unitsMeters$, and a surface coverage completeness of $\symCoverageRatio = 100\%$  was consistently achieved at each evaluated map resolution. 

\begin{figure*}[!htbp]
\captionsetup{justification=centering}
\centering
    \subfloat[Apt., $\symMapRes = 0.2 \unitsMeters$ \label{f:performanceResultsCombined:flat:voxelCounts}]{
		\centering
		\includegraphics[width=\imWidth,trim={1cm 0 0 0},clip]{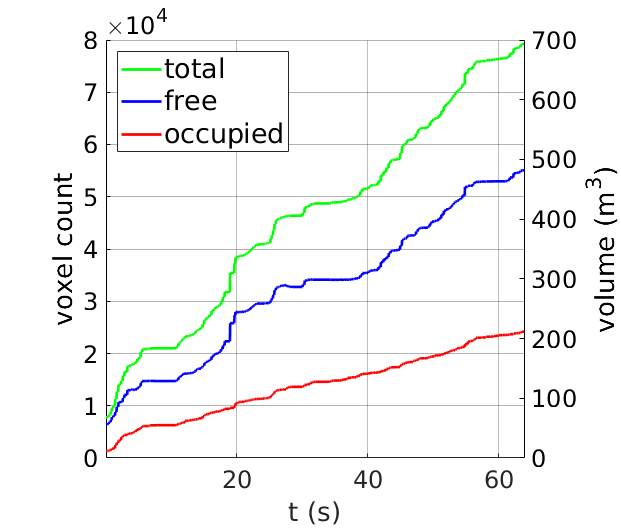}
	}
	\subfloat[Maze, $\symMapRes = 0.2 \unitsMeters$.\label{f:performanceResultsCombined:maze:voxelCounts}]{
		\centering
		\includegraphics[width=\imWidth,trim={1cm 0 0 0},clip]{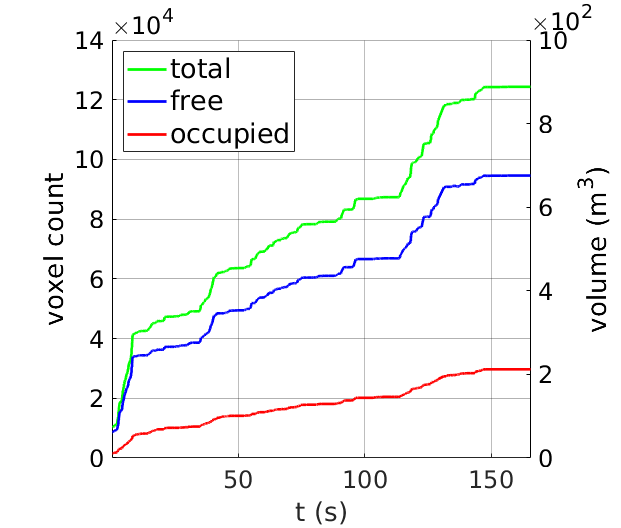}
	}
	\subfloat[Ind. Plant, $\symMapRes = 0.4 \unitsMeters$. \label{f:performanceResultsCombined:plant:voxelCounts}]{
		\centering
		\includegraphics[width=\imWidth,trim={1cm 0 0 0},clip]{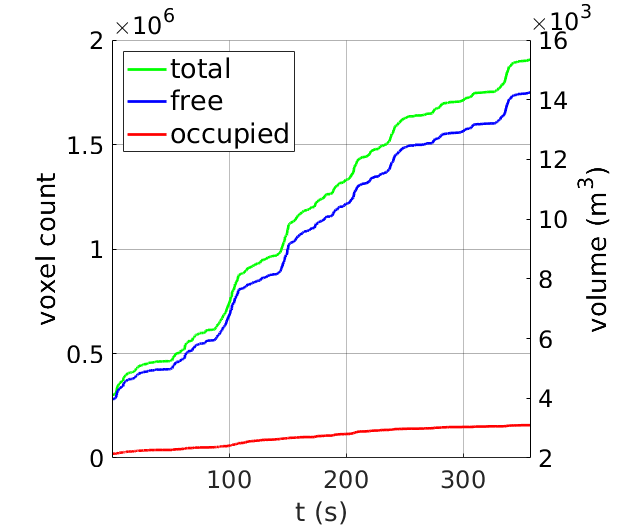}
	}
    \subfloat[Warehouse, $\symMapRes = 0.4 \unitsMeters$.\label{f:performanceResultsCombined:warehouse:voxelCounts}]{
		\centering
		\includegraphics[width=\imWidth,trim={1cm 0 0 0},clip]{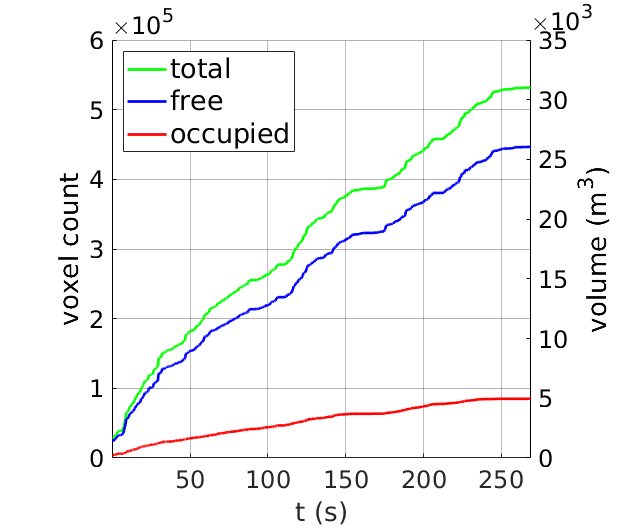}
	}
    \hfill \\
    \subfloat[Apt., $\symNodeDensityFactor = 17.1, \symEdgeDensityFactor = 0.58$.\label{f:results:flat:it_apn_size}]{
		\centering
		\includegraphics[width=\imWidth,trim={3mm 0 0 0},clip]{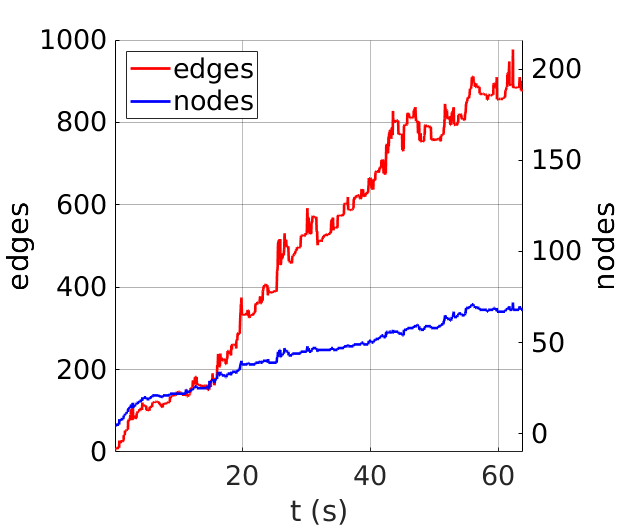}
	}
	\subfloat[Maze, $\symNodeDensityFactor = 16.0, \symEdgeDensityFactor = 0.20$.\label{f:results:maze:it_apn_size}]{
		\centering
		\includegraphics[width=\imWidth,trim={3mm 0 0 0},clip]{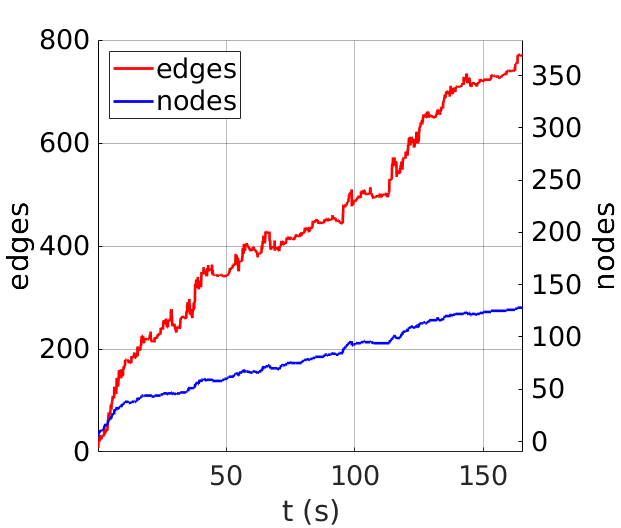}
	}
	\subfloat[Ind. Plant, $\symNodeDensityFactor = 3.8, \symEdgeDensityFactor = 0.25$. \label{f:results:plant:it_apn_size}]{
		\centering
		\includegraphics[width=\imWidth,trim={3mm 0 0 0},clip]{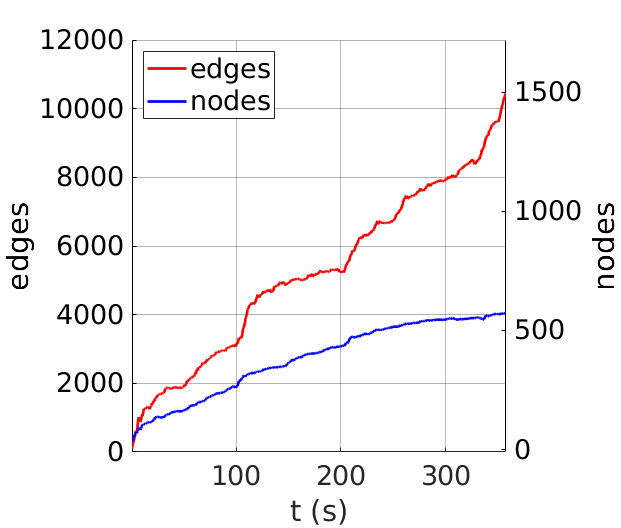}
	}
    \subfloat[Warehouse, $\symNodeDensityFactor = 2.7, \symEdgeDensityFactor = 0.42$.\label{f:results:warehouse:it_apn_size}]{
		\centering
		\includegraphics[width=\imWidth,trim={3mm 0 0 0},clip]{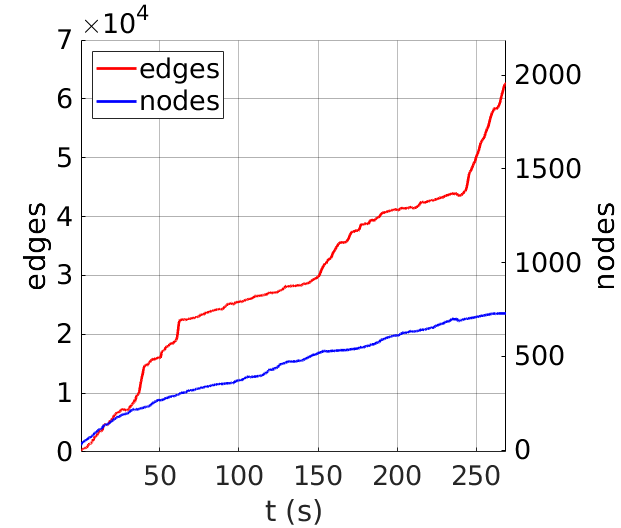}
	}
	
\caption{Representative results of the exploration progress over time. (a) - (d): explored map in terms of total voxels and their volume. (e) - (h): corresponding APN size in terms of its nodes (red) and edges (blue), with the respective node density ($\symNodeDensityFactor$) and edge density ($\symEdgeDensityFactor$).
}
\label{f:performanceResultsCombined}
\end{figure*}

Fig. \ref{f:performanceResultsCombined:flat:voxelCounts} shows an example of the explored map volume over time using resolution $0.2 \unitsMeters$ for reference. The surface coverage rate $\symSurfaceExploreRate$ was $1.5\CubicMetersPerSec$ and $2.6\CubicMetersPerSec$ for the respective map resolutions of $0.1\unitsMeters$ and $0.2\unitsMeters$. Since there are multiple dead-end regions for this scenario, some amount of backtracking is unavoidable, where the effects of backtracking correspond to the periods in Fig. \ref{f:performanceResultsCombined:flat:voxelCounts} where the map growth briefly stagnates (e.g. around the 30s timestamp). 

\assign{\imWidth}{0.236\textwidth}


The size growth of the APN over time shown in Fig. \ref{f:results:flat:it_apn_size}. Compared to the map scale in Fig. \ref{f:performanceResultsCombined:flat:voxelCounts}, the APN is significantly smaller and its growth over time is non-monotonic due to iterative pruning and refinements. The final state of the APN roadmap is shown in Fig. \ref{f:results:apartment:apn}, which can be seen to expand throughout the reachable free-space at a sufficient density for planning and navigation.

\begin{figure*}[!htbp]
\centering
\captionsetup{justification=centering}
	\subfloat[Apt. \label{f:computeTimesCombined:flat}]{
		\includegraphics[width=\imWidth]{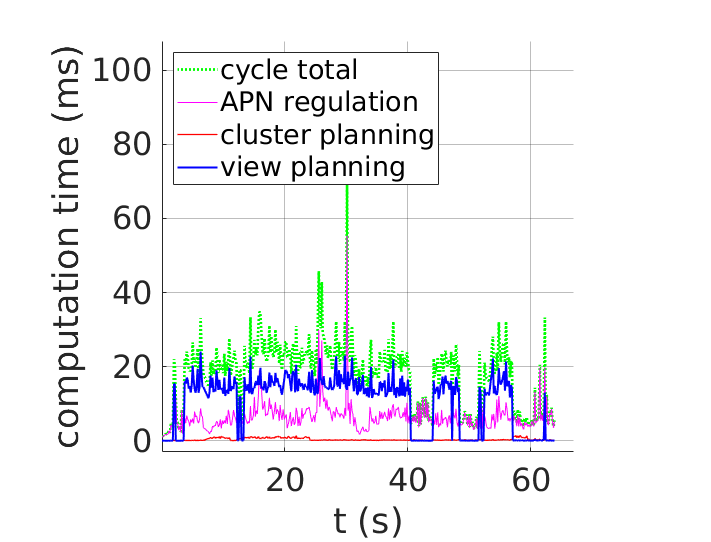}
	}
 	\subfloat[Maze  \label{f:computeTimesCombined:maze} ]{
		\includegraphics[width=\imWidth]{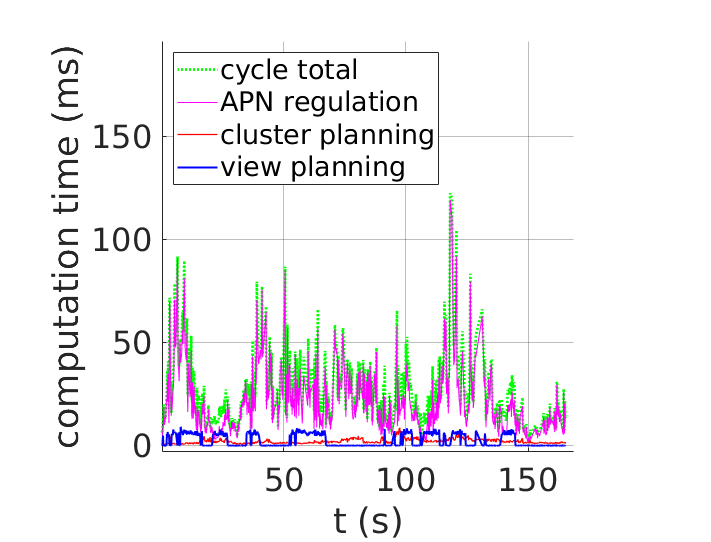}
	}
	\subfloat[Ind. Plant  \label{f:computeTimesCombined:plant} ]{
		\includegraphics[width=\imWidth]{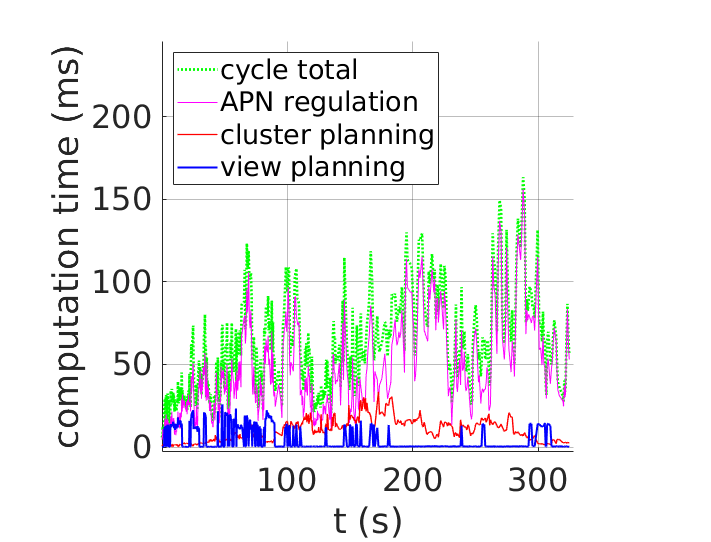}
	}
	\subfloat[Warehouse  \label{f:computeTimesCombined:warehouse} ]{ 
		\includegraphics[width=\imWidth]{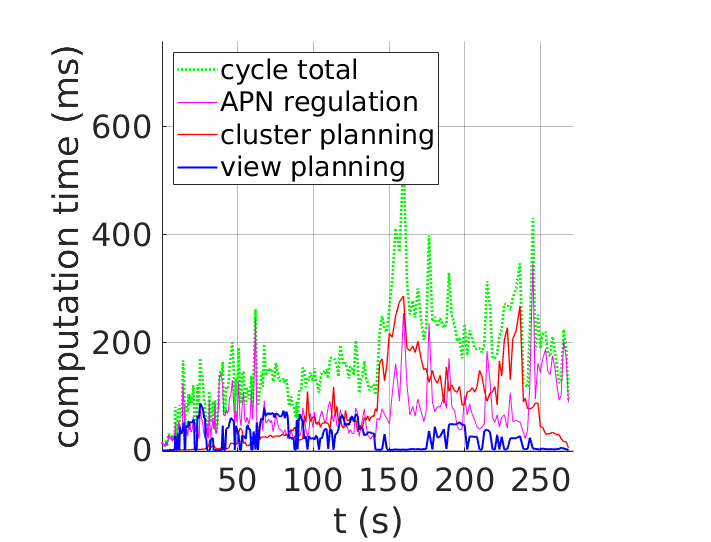}
	}
	\hfill
	\subfloat[Apt.  \label{f:processTime:boxplot:flat} ]{ 
		\includegraphics[width=\imWidth, trim={0 0 1cm 0 },clip ]{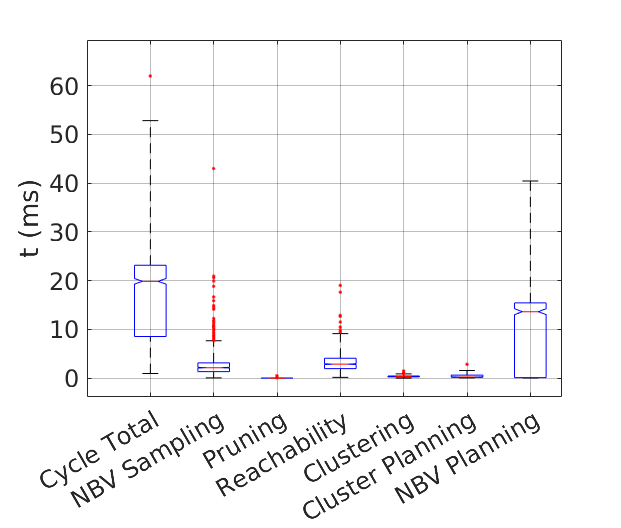}
	}
	\subfloat[Maze  \label{f:processTime:boxplot:maze} ]{
		\includegraphics[width=\imWidth, trim={0 0 1cm 0 },clip ]{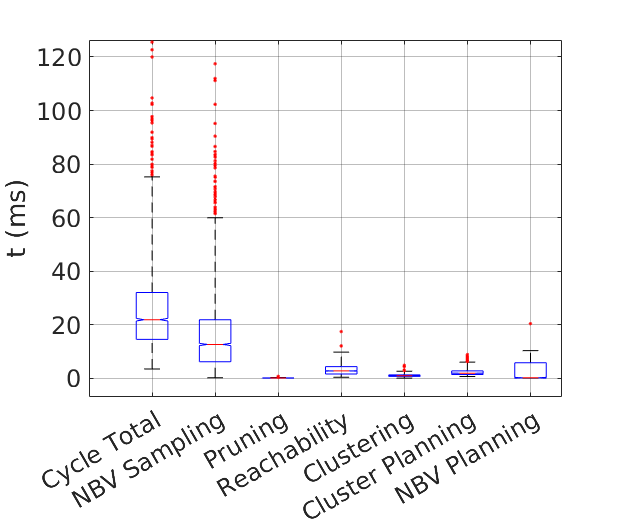}
	}
	\subfloat[Ind. Plant  \label{f:processTime:boxplot:plant} ]{
		\includegraphics[width=\imWidth, trim={0 0 1cm 0 },clip ]{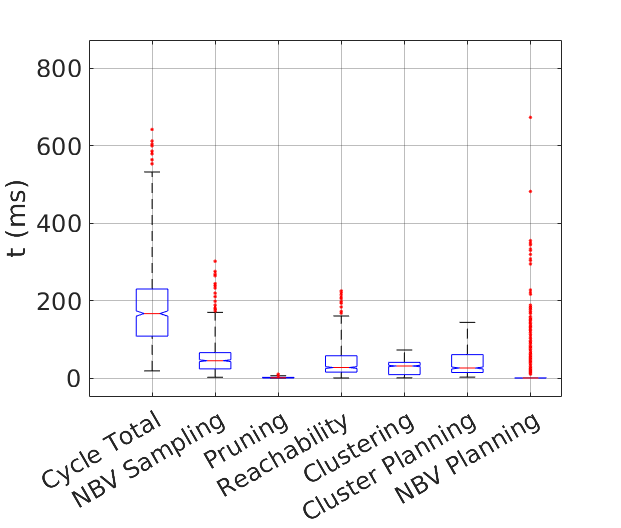}
	}
	\subfloat[Warehouse  \label{f:processTime:boxplot:warehouse} ]{
		\includegraphics[width=\imWidth, trim={0 0 1cm 0 },clip ]{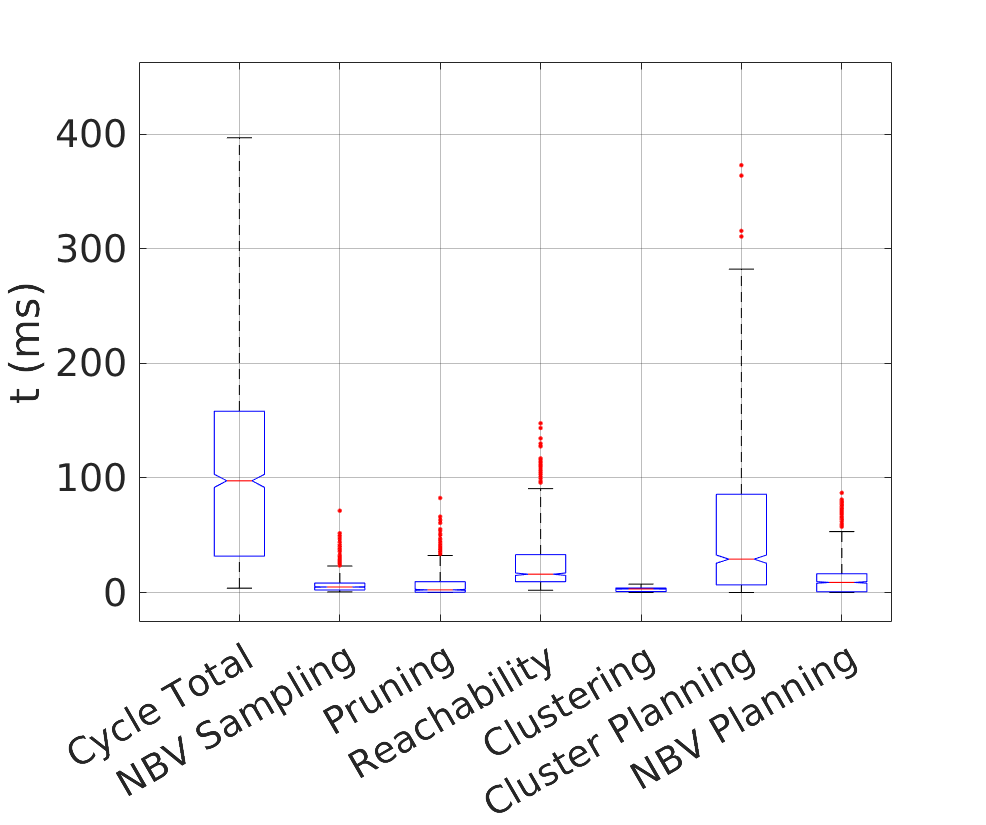}
	}
\caption{Timing performance for each exploration scenario. (a)-(d): depict the processing time taken per cycle. (e)-(h): display the median statistical boxplot of the DFR and planning computation times per cycle.}
\label{f:results:timingPerformance}
\end{figure*}

Fig. \ref{f:computeTimesCombined:flat} shows representative results of the computation times per cycle, using map resolution 0.2m as reference. The time taken for DFR remains fairly consistent over time despite the increasing map size. This demonstrates the effectiveness of the difference-aware update procedures at constraining the complexity as the map grows. A statistical boxplot of the respective procedures executed per cycle is shown in Fig. \ref{f:processTime:boxplot:flat}. The majority of computation time per cycle was spent on view planning, which had a median value of $13.6 \mSec$. The time spent on global cluster planning was negligible due to the relatively small size and complexity of this environment. The APN contained an average of only $1.2$ clusters, resulting in a trivial instance of cluster sequence optimization. The computation times for all differential regulation procedures were minimal compared to planning, given the relatively simple environment.

The time performance with the compared methods is summarized in Table \ref{t:timingPerformance:combined}. At the lowest map resolution of $0.4 \unitsMeters$, the APN-P achieved an average total exploration time of $\symRunTime = \apnpRes{apt}{totalTime}{0.4}\Sec \pm 4.3\Sec$, and average computation time per iteration of $\symCycleTime = \apnpRes{apt}{cycleTime}{0.4}\mSec$. Using a map resolution of 0.2m, the average exploration time was $\apnpRes{apt}{totalTime}{0.2}\Sec$ with $\apnpRes{apt}{cycleTime}{0.2}\mSec$ per cycle. At the highest map resolution of $0.1 \unitsMeters$, the average exploration time was $\apnpRes{apt}{totalTime}{0.1}\Sec $ with $ \apnpRes{apt}{cycleTime}{0.1}\mSec $ per cycle.

The RH-NBV approach required the highest total exploration time of 501.9s, with an average computation time per iteration of 153ms. For AEP, the total exploration time for each resolution was reported to take approximately 200s on average (exact quantities were not specified), with an average computation time per iteration of 98ms. FFI reported the fastest exploration time of the compared methods, with a total time of 80s and 151s for the respective map resolutions 0.4m and 0.1m. It should be noted that this approach was terminated once 95\% exploration was reached, rather than full coverage.

The APN-P performance demonstrated a significant improvement over the compared state-of-the-art implementations in terms of both total exploration time and per-iteration computation times. Compared to FFI, APN-P achieved complete coverage while the exploration time was reduced by $34\%$ using resolution $0.4 \unitsMeters$, and $54\%$ using resolution $0.1\unitsMeters$. Additionally, the percent improvement between resolutions indicates better scalability to higher resolution mapping.

\assign{\imWidth}{0.236\textwidth}
\unassign{\imWidth}

\renewcommand{\arraystretch}{1.4}
\begin{table*}[htbp]
\footnotesize
\caption{Time performance comparison in terms of total exploration runtime $\symRunTime$ and computation time per cycle $\symCycleTime$, averaged over 10 runs.}
\label{t:timingPerformance:combined}%
\centering
\makebox[0.97\textwidth][c]{
\begin{tabular}{cc|cc|cc|cc|cc|cc}
\toprule
    &  
    & \multicolumn{2}{c|}{\textbf{APN-P}} 
    & \multicolumn{2}{c|}{\textbf{FFI}  } 
    & \multicolumn{2}{c|}{\textbf{AEP}  } 
    & \multicolumn{2}{c|}{\textbf{RH-NBVP}  } 
    & \multicolumn{2}{c}{\textbf{Rapid}  } 
    \\
\midrule \midrule
\textbf{Scenario} & $\symMapRes[\unitsMeters]$ 
& $\symRunTime[\unitsSeconds]$ & $\symCycleTime[\unitsMillisec]$
& $\symRunTime[\unitsSeconds]$ &  $\symCycleTime[\unitsMillisec]$
& $\symRunTime[\unitsSeconds]$ &  $\symCycleTime[\unitsMillisec]$
& $\symRunTime$[\unitsSeconds] &  $\symCycleTime[\unitsMillisec]$ 
& $\symRunTime$[\unitsSeconds] &  $\symCycleTime[\unitsMillisec]$ 
\\
\midrule
\multirow{3}{*}{ \textbf{Apt.} } 
	& 0.4 
        & \apnpRes{apt}{totalTime}{0.4} & \apnpRes{apt}{cycleTime}{0.4}
		& 80   & $122 \pm 36$ 
		& ~200  & 92 
		& 501.9 & 153
		& - & - 
		\\
    & 0.2 
        & \apnpRes{apt}{totalTime}{0.2} & \apnpRes{apt}{cycleTime}{0.2}
		& -   & $156 \pm 109$
		& ~200  & - 
		& - & - 
		& - & - 
		\\
    & 0.1 
        & \apnpRes{apt}{totalTime}{0.1} & \apnpRes{apt}{cycleTime}{0.1}
		& 151   & $68 \pm 27$
		& ~200  & 129
		& - & - 
		& - & - 
		\\
    \hline
\multirow{2}{*}{\textbf{Maze} } 
	& 0.2 
        & \apnpRes{maze}{totalTime}{0.2} & \apnpRes{maze}{cycleTime}{0.2}
		& 177   & $155 \pm 71$ 
		& - & - 
		& - & - 
		& - & - 
		\\ 
    & 0.1 
        & \apnpRes{maze}{totalTime}{0.1} & \apnpRes{maze}{cycleTime}{0.1}
		& 330  & $238 \pm 80$
		& - & - 
		& - & - 
		& - & - 
		\\ 
    \hline
\textbf{Ind. Plant}
	& 0.2 
        & $\apnpRes{plant}{totalTime}{0.2}$ & $\apnpRes{plant}{cycleTime}{0.2} \pm 113.4 $
		& $\plantTotalTimeFFI$ & $\plantCycleTimeFFI$
		& $\plantTotalTimeAEP$ & $\plantCycleTimeAEP$
		& $\plantTotalTimeNBVP$ & $\plantCycleTimeNBVP$
		& $\plantTotalTimeRapid$ & $\plantCycleTimeRapid$
		\\ 
    \hline
\textbf{Warehouse} 
	& 0.4 
        & \apnpRes{warehouse}{totalTime}{0.4} & $\apnpRes{warehouse}{cycleTime}{0.4} \pm 84.4$
		& - & -
		& - & -
		& - & -
		& - & -
		\\ 
    \hline

\bottomrule
\end{tabular}
}
\end{table*}%
\renewcommand{\arraystretch}{1}

\subsection{Maze-like Scenario}
A maze-like environment is presented in Fig. \ref{f:worldScenarios:maze:worldModel} with the dimensions of $20 \times 20 \times 2.5 (\unitsMetersCubed$). This scenario was tested using map resolutions of $0.1 \unitsMeters$ and $0.2 \unitsMeters$; higher resolutions were not evaluated since there are narrow passageways that require lower resolutions to admit collision-free paths (as also noted in \cite{dai2020fast}). This scenario was primarily compared against FFI, as this scenario was not evaluated in the original works of the other approaches.

\begin{figure}[tbp]
\centering
\captionsetup{justification=centering}
\subfloat[Reconstructed map and exploration path. \label{f:results:maze:map}]{ 
	\includegraphics[width=0.7\columnwidth]{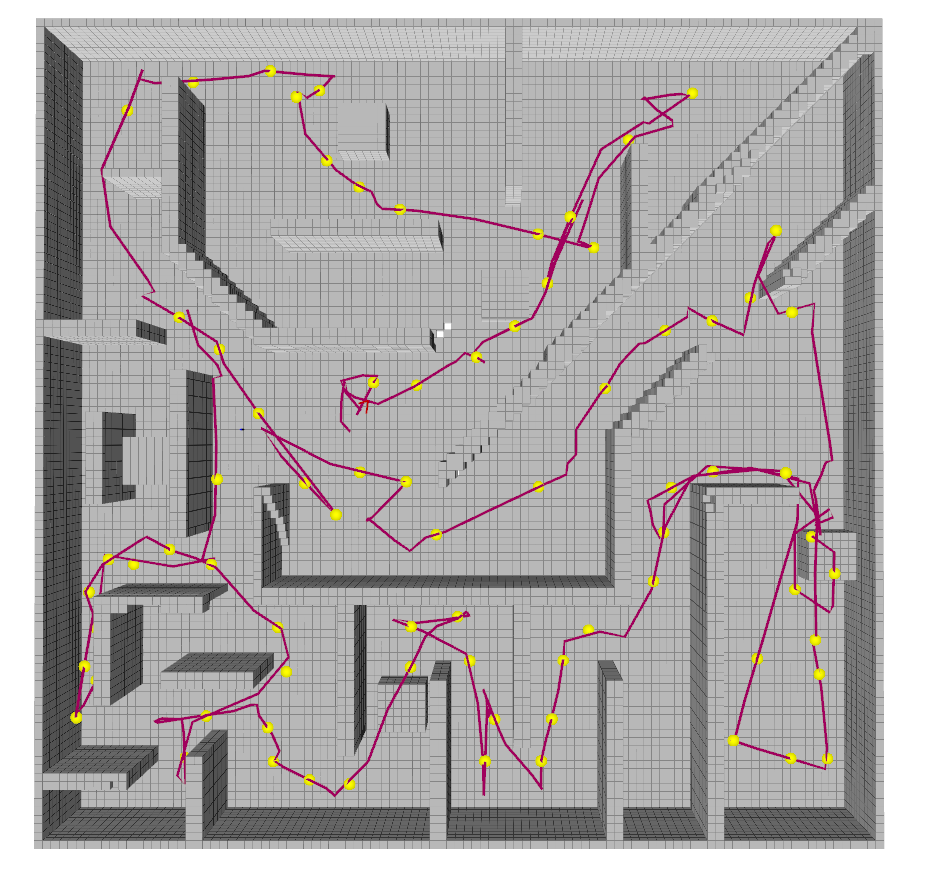}
    	\centering
}
\hfill
\subfloat[APN reachability roadmap. \label{f:results:maze:apn}]{ 
	\includegraphics[width=0.7\columnwidth]{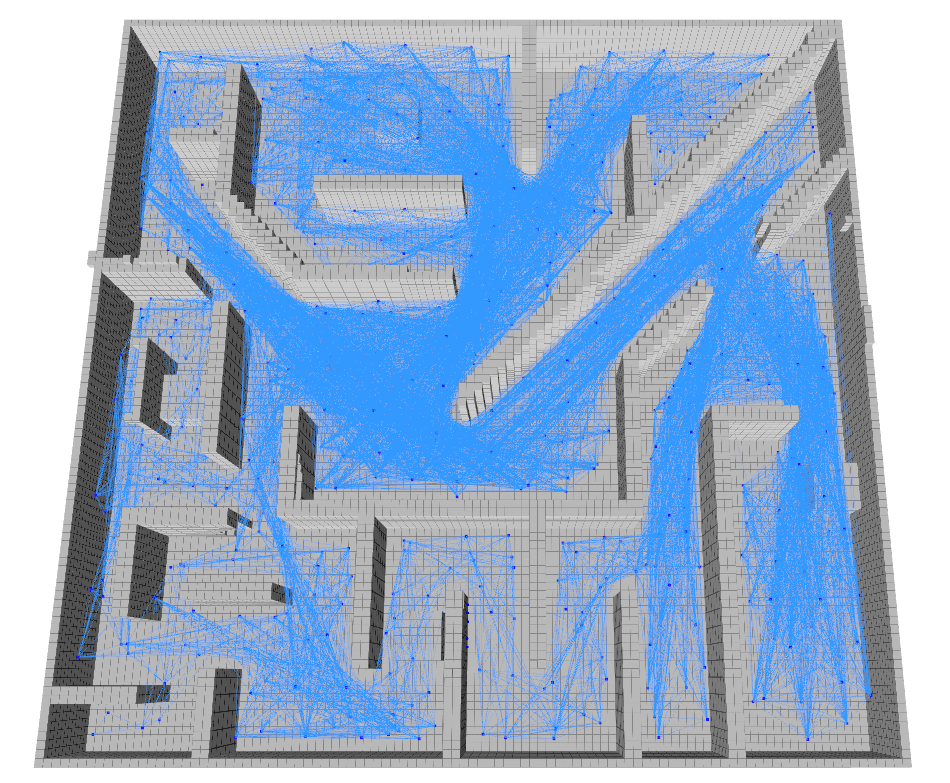}
	\centering
}
\caption{Exploration results for the Maze Scenario. (a): The explored path is plotted in red, with intermediate keyframe configurations represented by yellow points. (b): The APN nodes and edges overlayed in blue.}
\label{f:results:maze}
\end{figure}

A representative example of the mapped environment after exploration is shown in Fig. \ref{f:results:maze:map} with the executed exploration path overlayed in red. The path shows that very few redundant motions were executed and progresses smoothly throughout the maze passages, with an average total path length of $208.9 \unitsMeters$.

Fig. \ref{f:performanceResultsCombined:maze:voxelCounts} shows the map construction over time. An average coverage value of $\symCoverageRatio = 100 \%$ was reached at each map resolution, and the surface coverage rate $\symSurfaceExploreRate$ was $0.5\CubicMetersPerSec$ and $1.4\CubicMetersPerSec$ for the respective map resolutions of $0.1 \unitsMeters$ and $0.2 \unitsMeters$. The APN size growth over time was plotted in Fig. \ref{f:results:maze:it_apn_size}, and visualized in Fig. \ref{f:results:maze:apn}. The average node density per $100 \unitsMetersCubed$ was $\symNodeDensityFactor = 16.0 \pm 1.3$, with an average edge density of $\symEdgeDensityFactor = 0.20 \pm 0.12$. 

The computation times per cycle are plotted in Fig. \ref{f:computeTimesCombined:maze} with a statistical analysis of the computation time taken per procedure shown in Fig. \ref{f:processTime:boxplot:maze}. For this scenario, most of the computation time went towards APN regulation, with coverage view sampling requiring the most time of $15.8\mSec$ due to the prevalence of obstacles and occlusions. Despite the high obstacle density, the computation times for reachability updates remained relatively small, while still maintaining sufficient node and edge densities to facilitate planning. This demonstrates the effectiveness of the local difference-awareness and efficient data caching strategies that minimize wasteful or redundant processing. 

Table \ref{t:timingPerformance:combined} summarizes the exploration efficiency of the compared approaches with respect to total exploration time and computation time per cycle. Note that as previously mentioned, exploration time for FFI was reported when $95\%$ coverage was achieved, rather than $100\%$. The APN-P completed the exploration with $100\%$ coverage in an average time of $ \apnpRes{maze}{totalTime}{0.2} \Sec$ and $ \apnpRes{maze}{totalTime}{0.1} \Sec$ for map resolutions $0.2$m and $0.1$m, respectively. These are significant improvements over the results of FFI, while the processing time per cycle was also reduced by around $80\%$ and had much less variability. Additionally, the total exploration time for FFI increased by $86\%$ between the two map resolutions, while the respective increase for the APN-P was $45\%$. This further demonstrates the performance scalability for higher mapping resolutions using larger and more complex environments.

\subsection{Industrial Plant Scenario}

The Industrial Plant scenario shown in Fig. \ref{f:worldScenarios:plant:worldModel} is an outdoor environment based on the Gazebo Powerplant model, truncated to the approximate dimensions of $33 \times 31 \times 26 (\unitsMetersCubed$). It represents both a large-scale and complex exploration task due to intricate structural geometries with many auto-occlusions. It was tested using a map resolution of $0.2 \unitsMeters$ and maximum velocity of $2.5 \unitsVelocity$, consistent with the compared approaches.

\begin{figure}[!btp]
\centering
\captionsetup{justification=centering}
\subfloat[Reconstructed map (colorized by voxel height). \label{f:results:plant:map}]{ 
	\includegraphics[width=0.75\columnwidth]{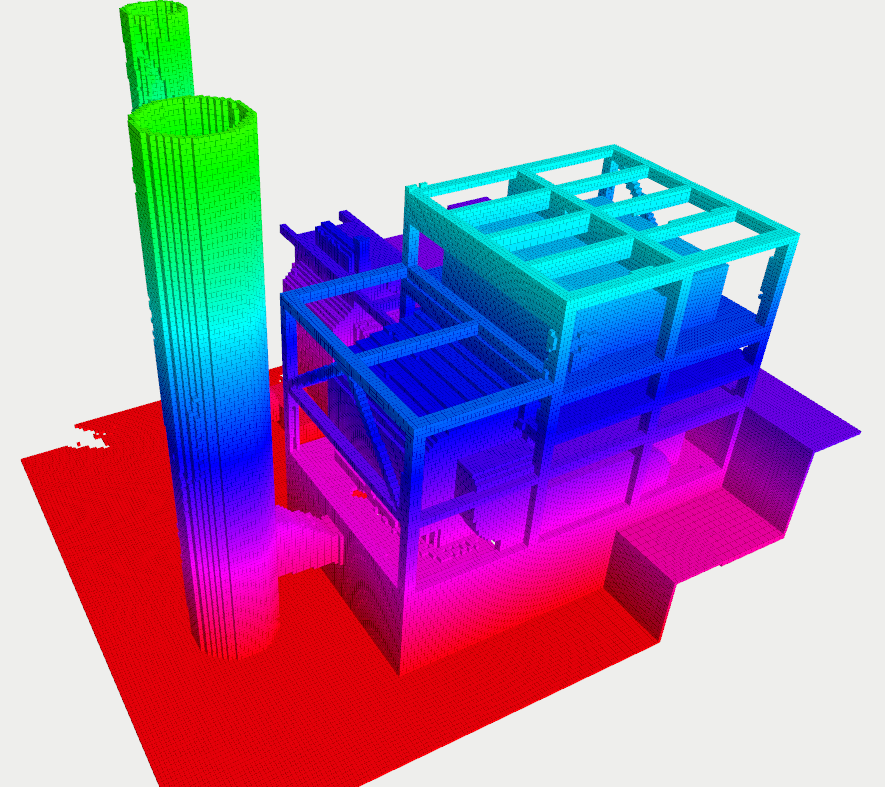}
    \centering
}
\\
\subfloat[APN reachability roadmap. \label{f:results:plant:apn}]{ 
	\includegraphics[width=0.75\columnwidth]{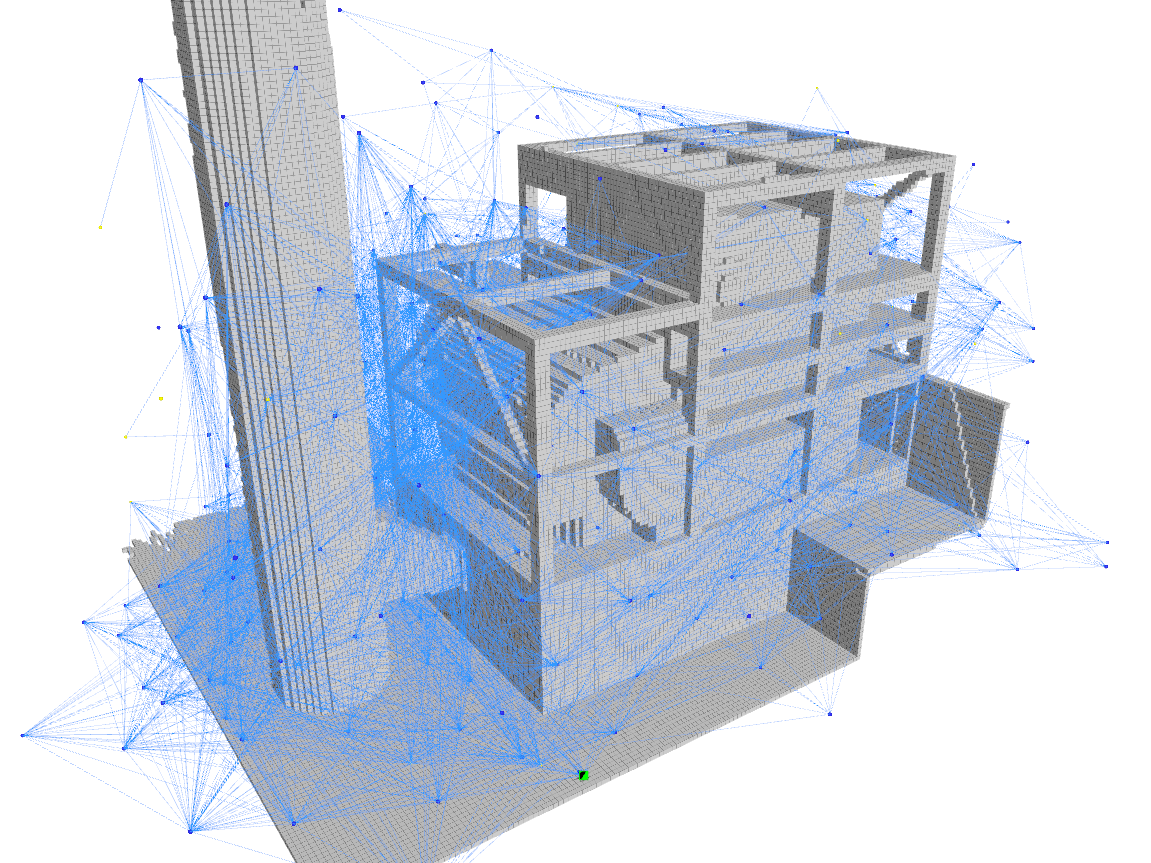}
    \centering
}
\caption{Exploration results of the Industrial Plant scenario.}
\label{f:results:plant}
\end{figure}

An example of the explored map is shown in Fig. \ref{f:results:plant:map}, with the explored volume over time plotted in Fig. \ref{f:performanceResultsCombined:plant:voxelCounts}. A high surface coverage rate of $\symSurfaceExploreRate = 3.2\CubicMetersPerSec$ was achieved, which was consistently maintained as shown in Fig. \ref{f:performanceResultsCombined:plant:voxelCounts}. The average total coverage was $\apnpRes{plant}{coverageRatio}{0.4}$, due to a few small regions with high surrounding occlusions, where coverage sampling failed to find a feasible viewpose. This could be overcome by selecting more aggressive sampling parameters, which was not done for these tests for parameter consistency between scenarios. 

The APN size over time is plotted in Fig. \ref{f:results:plant:it_apn_size}, with the final roadmap structure visualized in Fig. \ref{f:results:plant:apn}. The average node and edge density were $\symNodeDensityFactor = 3.8$ and $\symEdgeDensityFactor = 0.25$, respectively. By visual inspection of Fig. \ref{f:results:plant:apn}, the extent and density of the network appear to provide good coverage throughout the map.

The processing time per cycle is displayed in Fig. \ref{f:computeTimesCombined:plant}, with a statistical boxplot of the time taken by each subroutine shown in Fig. \ref{f:processTime:boxplot:plant}. Traversal edge maximization required the most computation time during differential regulation with an average of $67.2\mSec$ due to the large scale and the amount of empty-space surrounding the structures which is initially unknown. Unknown edges are repeatedly checked for collision checks until they can be determined as either completely free, or having an occupied collision after which they are suppressed. The processing time spent on planning was well-balanced between the hierarchical layers.

A comparison of the timing results to the baseline approaches is shown in Table \ref{t:timingPerformance:combined}. We note that this environment was not originally tested by the authors of RH-NBVP; instead, the corresponding value of $2104\Sec$ was obtained from the comparative analysis performed by \cite{cieslewski2017rapid}.

Rapid performed with the fastest total exploration time among the compared methods, taking $582\Sec$ with an average total distance of $728\Meters$. This was also the only approach able to finish exploration within the rated time limit $T_{max}$ of $840\Sec$. This could be explained because this approach takes advantage of the large amount of free space to maintain high velocity, which helps to offset the diminished efficiency from greedy planning. However, this also has the effect of frequently leaving regions that have only been partially mapped. Coverage gaps can frequently occur that require large redundant paths to revisit, or otherwise reduce the completeness of the final map depending on the specific termination criteria. 

Additionally, the authors of Rapid note that their implementation can spend a significant amount of time computing paths over large distances (up to 10 seconds) using Dijkstras algorithm over the map. These computation times were omitted from the reported total exploration time to focus evaluation only on the quality of their flight behavior. Even without this consideration, the APN-P was still able to reach complete exploration around $65\%$ faster on average with a decrease in distance traveled of around $10\%$. This also highlights the importance of the APN efficiency to prevent such high computation times from occurring in practice.

APN-P exhibited significantly better performance than all compared methods, requiring an average total exploration time of only $\apnpRes{plant}{totalTime}{0.2}\Sec$, with each cycle requiring an average of $\apnpRes{plant}{cycleTime}{0.2}\mSec$. The average total distance traveled was $\apnpRes{plant}{totalDist}{0.2}\Meters$, with mean velocity of $\apnpRes{plant}{AvgVel}{0.2}\unitsVelocity$. The MAV was able to maintain higher velocities due to the fast cycle times, which enabled the system to quickly react to the changing spatial map and re-plan its exploration path. Often the information gain of the current NBV goal gets fully observed as the MAV gets closer which can be quickly reflected within the network, allowing it to maintain its momentum by not needing to completely stop at each goal.

To evaluate how the larger size of this scenario correlates to the processing time per cycle, the Ind. Plant was additionally evaluated against the Maze scenario. To enable more consistent comparison, the map resolution was kept at $0.2 \unitsMeters$, and the maximum velocity and sensor parameters were assigned the values used for the Maze as indicated in Table \ref{t:apartment:params}. The resulting cycle processing time for the Ind. Plant decreased by around $58 \%$, with each cycle taking an average of $\symCycleTime = 78.7 \unitsMillisec$. Within each cycle, DFR required $46.4 \unitsMillisec$ and planning required $32.3 \unitsMillisec$. 

The effects of map resolution were analyzed by a testing the timing performance using map resolution of $0.4 \unitsMeters$. This resulted in a significant decrease in the cycle processing time, which was reduced to $\symCycleTime = 30.9 \unitsMillisec$, and the total exploration time was reduced to $\symRunTime = 220.2 \unitsSeconds$. This indicates the increased cycle processing time at resolution $0.2$ were primarily due to the increased resolution, rather than the larger environment size directly.

\subsection{Warehouse Scenario}

\assign{\valMaxVel}{3.0}
\assign{\valSensorRange}{9}
\assign{\valVFov}{75\degree }
\assign{\valHFov}{115\degree }
\assign{\valSafetyRadius}{ 1.0 }

The Warehouse scenario is a large-scale indoor environment with the approximate dimensions $90 \times 30 \times 15 \ (\unitsMetersCubed$), shown in Fig. \ref{f:worldScenarios:warehouse:worldModel} with its exterior shown on the left, and the interior structures shown on the right. The models exterior structure was derived from the Powerplant model available from the Gazebo model library, while the interior was modified by adding a various geometric features and structures to create a more intricate environment for exploration. Since this was a custom built model, the APN-P was evaluated independently as comparative results were unavailable.

Due to the larger scale of this scenario, the mapping resolution was set to $0.4 \unitsMeters$, and the maximum velocity was increased to $\valMaxVel \unitsVelocity$. The sensing parameters were also increased using a maximum range of $\valSensorRange \unitsMeters$, with FoV $(\valVFov, \valHFov)$. The larger sensor view volume results in more information being added to the map per scan and the higher maximum velocity results in more scans being integrated between cycles, both resulting in more changed data to process per cycle. This scenario was also used to analyze variations of the clustering parameters $\clusteringMinPts$ and $\clusteringEps$, which are indicated in Table \ref{t:warehouse:timingPerformance}. Unless otherwise noted, these parameters were set to $\clusteringMinPts = 4$ and $\clusteringEps = 7.0$, consistent with the previous Industrial Plant evaluation.

\begin{figure}[!bt]
\centering
\captionsetup{justification=centering}
\includegraphics[width=0.95\columnwidth]{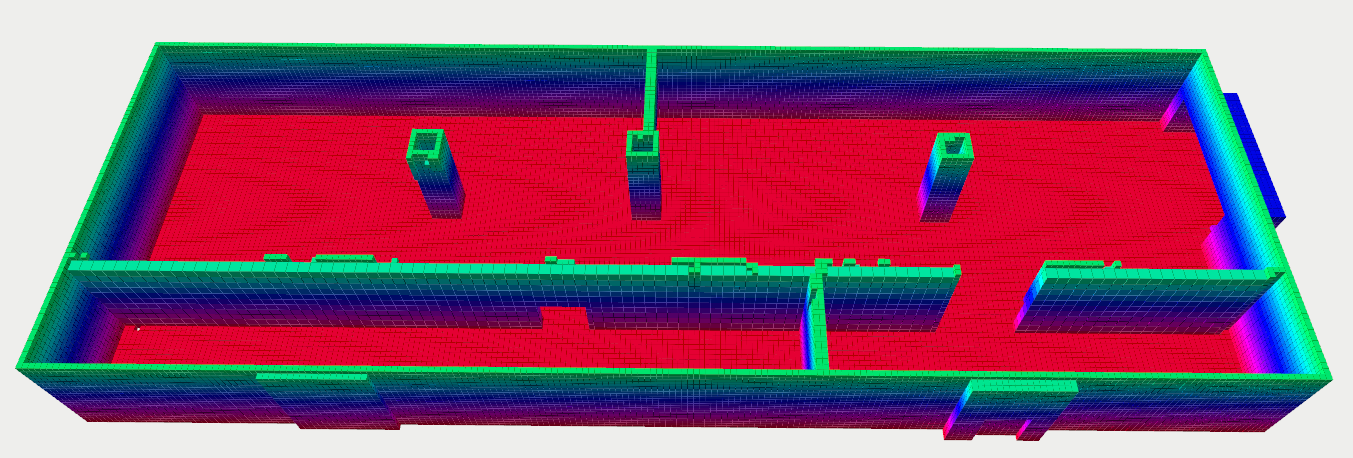}
\caption{The reconstructed map of the Warehouse scenario colorized by voxel height. The maximum height of displayed voxels is truncated for visual clarity.}
\label{f:results:warehouse:map}
\end{figure}

A representative example of the reconstructed map results shown in Fig. \ref{f:results:warehouse:map} and the explored map volume over time is shown in Fig. \ref{f:performanceResultsCombined:warehouse:voxelCounts}. A minimum coverage ratio of $\symCoverageRatio = \apnpRes{warehouse}{coverageRatio}{0.4}$ was achieved for all test configurations. The APN size growth is depicted in Fig. \ref{f:results:warehouse:it_apn_size}, which contained an average of $\apnpRes{warehouse}{AvgViewCount}{0.4}$ nodes and $\apnpRes{warehouse}{AvgEdgeCount}{0.4}$ edges, with an edge density factor of $0.374$.

The computation time per cycle is plotted in Fig. \ref{f:computeTimesCombined:warehouse} and summarized in Table \ref{t:timingPerformance:combined}. Similar to the Ind. Plant scenario, the time spent on APN regulation remains within a bounded range despite the increasing size of the map and APN. The exploration time performance results are summarized in Table \ref{t:timingPerformance:combined}, requiring an average exploration time of $ \symRunTime = \apnpRes{warehouse}{totalTime}{0.4} \Sec $ and average planning cycle time $\symCycleTime = \apnpRes{warehouse}{cycleTime}{0.4} \mSec$. A more detailed breakdown of the processing times per sub-procedure is shown in Fig. \ref{f:processTime:boxplot:warehouse}. 

Different clustering parameter variations were applied and the resulting time performance is summarized in Table \ref{t:warehouse:timingPerformance}. The average exploration time was not significantly changed between parameter variations, indicating the low sensitivity of these parameters. The primary effect of the variations was on the per-cycle computation time, though the differences were relatively minor. Using the values $\clusteringMinPts = 4$ and $\clusteringEps = 10.0$, the cycle time was nearly evenly distributed between differential regulation, $\symCycleTimeAPN$, and planning, $\symCycleTimePlanning$. The other parameter combinations increased the planning time, but only by a small amount.

\renewcommand{\arraystretch}{1.4}
\begin{table}[tbp]
    \footnotesize
    \caption{Timing performance for the Warehouse Scenario with to variations of the clustering parameters $\clusteringMinPts$ and $\clusteringEps$.}
    \label{t:warehouse:timingPerformance}%
    \centering
\begin{tabular}{| cc | cccc |}
\toprule
    $\clusteringMinPts$ & $\clusteringEps[\Meters]$ 
    & $\symRunTime[\Sec]$ & $\symCycleTime[\mSec]$ & $\symCycleTimeAPN[\mSec]$ & $\symCycleTimePlanning[\mSec]$ \\ 
\hline
4 & 7.0 
    & $\apnpRes{warehouse}{totalTime}{0.4}$  & $\apnpRes{warehouse}{cycleTime}{0.4}$ 
    & 47.4 & 78.9 \\
4 & 10.0    
    & $270.0$ & $109.7$ 
    & 55.3 & 54.4 \\ 
7 & 10.0    
    & $274.9$ & $128.7$ 
    & 49.1 & 79.7 \\
\bottomrule
\end{tabular}
\end{table}%
\renewcommand{\arraystretch}{1}

\subsection{Discussions} \label{sec:discussion}

The experimental results show that our approach has the ability to iteratively update the APN and replan the exploration path at an average rate of at least $20 \unitsHz$ for the two smaller scale scenarios (Apt. and Maze), and at least $5 \unitsHz$ for the larger scales (Industrial Plant and Warehouse). However, the difference between these cycle rates is not primarily due to the larger environment sizes. Instead, the larger sensor view volume and higher maximum velocities are the more significant factors, which result in a larger amount of map data for processing per cycle, but these factors are not directly related to the environment size. This helps to explain the scalability of our approach for larger environments.  

For the smaller environments, most of the planning time is spent on local view planning (see Fig. \ref{f:processTime:boxplot:flat} and \ref{f:processTime:boxplot:maze}), This is due to the relatively few clusters needed to partition the nodes, resulting in trivial cluster planning instances. However, planning directly over all NBVs can quickly become intractable as the map size increases, either resulting in unacceptably large processing times, or would otherwise require premature search termination that degrades the planning quality. 

The hierarchical planning strategy of APN-P helps to mitigate the complexity by keeping the problem size manageable. Furthermore, planning convergence is further accelerated by initializing each planning cycle from the partially optimized solution of the previous cycle. This reduces the need to introduce further problem simplifications or approximations that would decrease the planning quality. These effects are demonstrated by the results shown in Fig. \ref{f:computeTimesCombined:warehouse} and \ref{f:processTime:boxplot:warehouse}. The distributed planning time remains relatively low and does not exhibit continually increasing growth, despite the increasing size of the map and APN as shown in Fig. \ref{f:results:warehouse:it_apn_size}, \ref{f:performanceResultsCombined:warehouse:voxelCounts}. 

The frontier-guided information gain and sampling strategy of DFR provides an effective way to 
avoid the prohibitively high computation costs for analyzing information gain by the existing (compared) approaches and to balance processing time per cycle and update rates. This enables maximized coverage of the unknown map regions to be maintained at high update rates, providing the necessary knowledge needed for non-myopic planning.

\section{Conclusions and Future Work} \label{conclusions}

This paper has presented the Active Perception Network (APN), serving as a topological roadmap of the dynamically changing exploration state space, the differential regulation (DFR) update procedure that incrementally adapts the APN to the changing environment knowledge, and an exploration planner APN-P, which leverages the APN to find non-myopic exploration sequences through the APN. 

The results demonstrate the efficiency of DFR in performing  each cyclic update and its scalability with increasing map sizes. In comparison to several state-of-the-art approaches, the APN-P consistently demonstrated improved performance in terms of total exploration time and coverage completeness. The improved performance was achieved over a variety of different environments, both indoor and outdoor, with only minor parameter adjustments between them. We expect to make all implementations of the presented work available as open-source, including the full development framework it was built upon (briefly introduced in the Appendix).

Several areas of future work have been identified. An investigation of different clustering methods and their performance effects will provide insight toward future improvements. Methods to account for sensing and localization uncertainty using multi-objective optimization strategies are also being investigated, where the current processing performance provides an excellent baseline to account for the greater computational complexities inherent to these. This will make the approach more robust for practical use in GPS-denied environments. An ablation study and analysis of parameter sensitivity will help guide future developments that are generalizable to a wider range of environment scenarios and that reduce or eliminate the need for parameter tuning.

\bibliographystyle{IEEEtran}
\bibliography{references}

\appendix

\section{Appendix}
\subsection{Software framework and architecture}

A C++ application development framework,  which we refer to as Active Perception for Exploration, Mapping, and Planning (APEXMAP), was created in conjunction with the proposed approach and used to implement this work. The main goal of APEXMAP is to facilitate the rapid development, implementation, and evaluation of complex autonomous exploration and active perception systems for a variety of different robots, sensors, and applications. Its structure consists of a distributed collection of decoupled library packages, each intended to address a well-defined subproblem of active perception (e.g. mapping, sensing, visibility analysis) using generic and modular programming paradigms. These form the building blocks for the design of a complete system. While a detailed description of its complete structure is beyond the scope of this paper, we briefly introduce some of its key concepts.

Unlike other open-source software for exploration tasks, APEXMAP is not restricted to a particular exploration formulation or methodology. As such, the APN is represented as a modular graph model, where its structure is not limited to that presented in this paper, but can be reconfigured to meet the requirements of nearly any graph-based formulation. For example, it supports the ability to represent a single graph structure or composition of multiple subgraphs using either directed or undirected edges, and the type and quantity of attributes for its nodes and edges can be freely defined. These characteristics can also be extended to define any number of hypergraph structures. The low level data structures and data types are optimized using similar design patterns to the LEMON graph library \cite{dezsHo2011lemon}, which can achieve O(1) worst case amortized time complexity for all fundamental search and access operations, with modifying operations in linear worst case time complexity. 

Along similar lines, DFR and planning subsystems are modeled by a standardized front-end interface, while the back-end can be fully redefined using Inversion of Control (IoC) principles. This allows different update strategies to be freely implemented on a given APN configuration. 

An event-driven architecture (EDA) is used to provide inline difference-awareness throughout the system, and enable dynamic data structures critical for managing dynamically changing data to ensure its state remains consistent across a complex system. Combined with IoC principles, any data container or other structures can be configured to broadcast the event when data inserted, erased, or modified. Other data structures can then subscribe to these events and invoke a user-defined callback to handle them. 

These design patterns decouple the transfer of data between interfaces, allowing behaviors to be quickly and transparently changed throughout the system during its development and evaluation. They also prevents wasteful or redundant operations such as search and comparison, which may otherwise be needed to determine what data has been changed. An additional benefit is that many algorithms can be made more efficient by only processing incremental changes to its inputs, rather than reprocessing the full input set (e.g. maintaining the a list of connected components over the graph). The framework provides generic and flexible interfaces that enable these to be quickly implemented or modified.

\end{document}